\documentclass{bmcart}

\usepackage[english]{babel}
\usepackage{amsfonts}
\usepackage{amssymb}
\usepackage{amsmath}
\usepackage{booktabs}
\usepackage{mathrsfs}
\usepackage{dsfont}
\usepackage{bm}
\usepackage[T1]{fontenc}

\usepackage[hyphens]{url}
\usepackage[hypertexnames=false]{hyperref}
\usepackage{caption}
\usepackage{subcaption}

\usepackage[utf8]{inputenc} 
\usepackage{graphicx}

\startlocaldefs

\newcommand{\field}[1]{\mathbb{#1}}

\endlocaldefs

\begin{document}

\begin{frontmatter}

\begin{fmbox}
\dochead{Research}


\title{Unveiling the invisible - mathematical methods for restoring and interpreting illuminated manuscripts}


\author[
   addressref={aff1},                   
   email={luca.calatroni@polytechnique.edu}   
]{\inits{L}\fnm{Luca} \snm{Calatroni}}
\author[
   addressref={aff2},                   
   email={mariedautume@gmail.com}   
]{\inits{M}\fnm{Marie} \snm{d'Autume}}
\author[
   addressref={aff3},
   email={}
]{\inits{r}\fnm{Rob} \snm{Hocking}}
\author[
   addressref={aff5},
   email={sdp26@cam.ac.uk}
]{\inits{S}\fnm{Stella} \snm{Panayotova}}
\author[
   addressref={aff4},
   email={sp751@cam.ac.uk}
]{\inits{S}\fnm{Simone} \snm{Parisotto}}
\author[
   addressref={aff5},
   email={pr364@cam.ac.uk}
]{\inits{S}\fnm{Paola} \snm{Ricciardi}}
\author[
   addressref={aff3},
   corref={aff3},                       
   email={cbs31@cam.ac.uk}
]{\inits{CB}\fnm{Carola-Bibiane} \snm{Sch\"{o}nlieb}}


\address[id=aff1]{
  \orgname{CMAP, \'Ecole Polytechnique},        
  \street{Route de Saclay},                   %
  \postcode{91128}                            
  \city{Palaiseau},                               
  \cny{FR}                                    
}
\address[id=aff2]{
  \orgname{CMLA \'Ecole Normale Sup\'erieure Paris-Saclay},        
  \street{61 Avenue President Wilson},                   %
  \postcode{94230}                            
  \city{Cachan},                               
  \cny{FR}                                    
}
\address[id=aff3]{%
  \orgname{DAMTP, University of Cambridge},
  \street{Wilberforce Road},
  \postcode{CB3 0WA}
  \city{Cambridge},
  \cny{UK}
}
\address[id=aff4]{%
  \orgname{CCA, University of Cambridge},
  \street{Wilberforce Road},
  \postcode{CB3 0WA}
  \city{Cambridge},
  \cny{UK}
}
\address[id=aff5]{%
  \orgname{Fitzwilliam Museum, University of Cambridge},
  \street{Trumpington Street},
  \postcode{CB2 1RB}
  \city{Cambridge},
  \cny{UK}
}


\begin{artnotes}
\end{artnotes}

\end{fmbox}


\begin{abstractbox}

\begin{abstract} 
The last fifty years have seen an impressive development of mathematical methods for the analysis and processing of digital images, mostly in the context of photography, biomedical imaging and various forms of engineering. The arts have been mostly overlooked in this process, apart from a few exceptional works in the last ten years. With the rapid emergence of digitisation in the arts, however, the arts domain is becoming increasingly receptive to digital image processing methods and the importance of paying attention to this therefore increases. In this paper we discuss a range of mathematical methods for digital image restoration and digital visualisation for illuminated manuscripts. The latter provide an interesting opportunity for digital manipulation because they traditionally remain physically untouched. At the same time they also serve as an example for the possibilities mathematics and digital restoration offer as a generic and objective toolkit for the arts.

\end{abstract}


\begin{keyword}
\kwd{Sample}
\kwd{Mathematical image reconstruction}
\kwd{Image inpainting}
\kwd{Image osmosis}
\kwd{3D visualisation}
\end{keyword}


\end{abstractbox}
%

\end{frontmatter}



\section*{Content}
Text and results for this section, as per the individual journal's instructions for authors. 

\section{Introduction} \label{sec:intro}

The digital processing, analysis and archiving of databases and collections in the arts and humanities is becoming increasingly important. This is because of a myriad of possibilities that digitisation opens up that go well beyond the organisation and manipulation of the actual physical objects, allowing, for instance, the creation of digital databases that are searchable with respect to several parameters (keywords), the digital processing
and analysis of objects that are non-destructive to the original object, and the application of automated algorithms for sorting newly found objects into existing digital databases by classifying them into pre-defined
groups in the database. These possibilities go hand-in-hand with ever-growing advances in data science that are developing mathematical methodology for analysing and processing digital data. A large component of
digital data in the arts and humanities is composed of digital images. Despite many developments of mathematical image analysis methods in applications in biomedicine, the physical sciences and various forms of engineering, the arts and humanities have been, apart from only a few exceptions and more recent works of, for instance, Daubechies and co-authors \cite{polatkan2009detection,ruvzic2011virtual,anitha2013restoration,cornelis2013crack,pizurica2015digital,yin2016removing}, Fornasier \cite{fornasier2005fast,fornasier2007restoration}, Abry, Jaffard and co-authors \cite{Wendt2013a, abry2015multiscale}, and \cite{kirchner2018digitally,kirchner2018digitally2,kirchnerdigitally3,van2017learned,van2015toward}, have been mostly overlooked as an application in need of bespoke mathematical image analysis methods.

In this work we discuss a range of mathematical methods for automated digital restoration based on partial differential equations, exemplar-based image inpainting and osmosis filtering, and, by examples, their translation to the digital restoration and interpretation of illuminated manuscripts. The pre-sequel of this article is an article in the exhibition catalogue \cite{panayotova2016}.

Medieval and Renaissance illuminated manuscripts present a particular challenge, but also an opportunity to transform current understanding of European visual culture between the 6th and 16th century. Illuminated manuscripts are the largest and best preserved resource for the study of European painting before 1500. Nevertheless, the images in some manuscripts have been affected by wear-and-tear, degradation over time, iconoclasm, censorship or updating. Unlike the conservation of other painted artefacts, the conservation of illuminated manuscripts preserved in institutional collections is non-invasive, usually restricted to repairs of the binding and of torn parchment or paper, and rarely involves the consolidation of flaking pigments. Damaged areas in the images are never restored; pigment losses are never filled in; over-painting added on top is never removed to reveal the original images. This minimal approach is due largely to the fact that when compared to wall or easel paintings, the images in illuminated manuscripts are relatively small and their pigment layers are few and very delicate. It is not possible to remove over-painting without damaging or completely removing the original painting beneath. The removal of even the smallest sample or the restoration of even the smallest painted area would constitute a considerable change to the overall image.  
Virtual restoration is thus the only way to recover damaged illuminations, whether by infilling paint losses or by removing over-painted layers or indeed both. Bringing the images as close as possible to their original form would ensure both their accurate scholarly interpretation and their full appreciation by wider audiences. Damaged or inaccurately restored illuminations can lead to the exclusion of seminal works of art from academic debates or to incomplete and misleading interpretations of the dating, origin and artists involved. Preserving the current state of the illuminations in line with conservation ethics, faithful digital restoration would serve as a reliable surrogate for multiple reconstructions, enabling research, teaching and wider appreciation for manuscripts.

Reliable digital restoration for illuminated manuscripts requires a multi-disciplinary collaboration as the current work is based on. In what follows we discuss a range of new adaptive, semi-automated restoration methods that (a) reconstruct image-structures using partial differential equations \cite{bertalmio2000image,masnou1998level,Bornemann2007,Baatz2008,Baatz2009,Burger2009,CarolaBook}, (b) mimic the human-expert behaviour, using texture- and structure patches sampled from the intact part of the illuminated manuscript at hand and integrating them in exemplar-based inpainting approaches \cite{Criminisi2004,Arias2011}, (c) exploit infrared imaging data, correlating the visible image content with its traces in the hidden layers of paint \cite{weickert2013linear,panayotova2016}, and (d) create new 3D interpretations of illuminated manuscripts through a new 3D  conversion pipeline \cite{Guidefill}. 

\paragraph{Organisation.} 
In Section \ref{sec:illuminated_manuscripts} we propose a semi-supervised approach for the segmentation of damaged areas in manuscripts and for the retrieval of missing information within them via an exemplar-based image inpainting model. In section \ref{sec:osmosis} we consider the mathematical model of image osmosis to integrate super-painted visible image information on a manuscript with hidden infrared ones for looking through the layers of a restoration process. Finally, in Section \ref{sec:3D visualisation} we present a mathematical pipeline to convert a 2D painting into a 3D scene by means of the construction of an appropriate depth map.

\section{Retrieving missing contents via image inpainting}
\label{sec:illuminated_manuscripts}
The problem of image inpainting can be described as the task of filling in the damaged (or occluded) areas in an image $f$ defined on a rectangular domain $\Omega$ by transferring the information available in the visible areas to the damaged areas in the image. 
Over the last thirty years a large variety of mathematical models solving the image inpainting problem have been proposed, see, e.g., \cite{ChanShenBook,CarolaBook} for a review. 
In some of them, image information is transferred into the damaged areas (the so-called \emph{inpainting domain}, denoted by $D$ in the following) by using \emph{local} information only, i.e.\ by means of suitable diffusion and transport processes which interpolate image structures in the immediate vicinity of the boundary of $D$ inside the occluded region. 
Such techniques have been shown to be effective for the transfer of geometrical image structures, even in the presence of large damaged areas \cite{CarolaBook}. 
However, because of their local nature, such methods do not make use of the entire information contained in the intact image regions. 
In particular, such methods do not take into account image information located far apart of $D$ to estimate the missing contents. For such a reason, more sophisticated (non-local) mathematical models exploiting self-similarities in the whole image at hand have been proposed \cite{Criminisi2004,Arias2011,Newson2014}. 
Heuristically, they can be thought as copy and paste models where small patches inside $D$ are iteratively reconstructed by comparison with patches outside $D$ in terms of a suitable distance. 
These models have been proven to be impressively effective in a very large variety of applications and rendered computationally feasible in recent years with the well-known PatchMatch algorithm \cite{Barnes2009}.

The very first basis step of any inpainting algorithm consists in the localisation of the damaged areas. This is essentially an image segmentation problem which may be rendered very hard in the presence of fuzzy and irregular region boundaries and small scale objects.

In the following we describe a semi-supervised algorithm for the detection of the damaged areas in images with possibly large and non-homogeneous missing regions. This is then used  as a necessary initial step for the subsequent application of the exemplar-based non-local image inpainting model proposed in \cite{Newson2014} to the reconstruction of image contents in two highly-damaged images \ref{fig:illuminated manuscripts} of illuminated manuscripts.

\subsection{Description of the dataset} \label{sec:dataset manuscripts}
Our dataset is made by two manuscripts, reported in Figure \ref{fig:illuminated manuscripts}: William de Brailes, \emph{Last Judgement} (left) and \emph{Christ in Majesty with King David playing the harp} (right), Fitzwilliam Museum, MSS 330.iii and 330.v,\ England, Oxford, c.\ 1230-1250.
The images to be processed are very large \texttt{.tif} data of $4008\times 5344$ pixels and approximatively $47$MB each.

\subsection{A semi-supervised algorithm for the detection of the damaged areas}  \label{sec:segmentation domain}
In order to identify the damaged areas in the image, we propose in the following a combined algorithm where a classical binary segmentation model is used for the extraction of a small training region which serves as an input for a subsequent labelling algorithm which segments the whole inpainting domain in terms of appropriate intensity-based image features.
%
\subsubsection{Chan-Vese binary segmentation.}  \label{sec:chan-vese}
A binary image segmentation task consists in the partitioning of an image into two different regions characterised each by distinctive features. 
Typically, RGB intensity values are used to describe image contents and mathematical image segmentation methods compute the required segmented image as the minimiser of an appropriate functional where information on the magnitude of the gradient of the image are encoded in order to identify the regions (boundaries) where intensity values drastically change.  


If by simplicity one assumes that the image $f$ at hand can be approximated by a binary function $u$ so that 
\begin{equation}  \label{modelchanvese}
u(x) = \begin{cases} c_1, &\mbox{if } x\mbox{ is inside }C, \\
c_2, & \mbox{if } x \mbox{ is outside }C, \end{cases}
\end{equation}
where $C$ is a closed curve, a very well-known model computing the segmentation of $u$ in two classes is the Chan-Vese model presented in \cite{ChanVese2001}. For such modelling, the functional to minimise seeks the optimal $c_1$ and $c_2$, i.e.\ the optimal $u$ of the form \eqref{modelchanvese}, and contour $C$ to minimise
\begin{eqnarray}  \label{eq:chanvese}
\mathcal{F}(c_1,c_2,C):&=& \mu~ \text{Length}(C) + \nu~\text{Area}(int(C))\\ & &+  \notag  \lambda_1\sum_{x\in int(C)}  | f(x)-c_1 |^2 + \lambda_2\sum_{x\in ext(C)} |f(x)-c_2|^2 
\end{eqnarray}
where $\mu,\,\nu,\,\lambda_1,\,\lambda_2>0$ and $int(C),\, ext(C)$ denote the inner and the outer part of $C$, respectively.
In \eqref{eq:chanvese} the first term penalises the length of $C$, ensuring its regularity. 
The size of $C$ is controlled by the second term which is penalising the area in the interior of $C$, while the two other terms penalise the discrepancy between the fitting of the piecewise constant $u$ with the model \eqref{modelchanvese} and the given image $f$ in the interior and exterior of $C$, respectively. 
By computing the local minima of \eqref{eq:chanvese} one retrieves the optimal binary approximations $u$ of $f$.

Despite being very popular and widely used in applications, the Chan-Vese model and its extensions present intrinsic limitations. 
Firstly, the segmentation result is strongly dependent on the initialisation: in order to get a good result, the initial condition needs to be chosen within (or sufficiently close to) the domain one aims to segment. 
Secondly, due to the modelling assumption \eqref{modelchanvese}, the Chan-Vese model works well for images whose intensity is locally homogeneous. If this is not the case, the contour curve $C$ may evolve along image information different from the one we want to detect.  
Images with significant presence of texture, for instance, can exhibit similar problems. 
Furthermore, the model is very sensitive to the length and area parameters $\mu$ and $\nu$, which may make the segmentation of very small objects in the image very difficult. 

For our application, we make use of the Chan-Vese model to segment a sub-region $D_1$ included in $D$ out of the whole image and with a high level of precision \footnote{For our computation we used the inbuilt MATLAB Chan-Vese segmentation code.}. 
To do that, we ask the user (typically, an expert in the field) simply to click in few pixels of the inpainting domain to identity a candidate initial condition for the segmentation model \eqref{modelchanvese}, which is then run to segment the subregion $D_1$. 
In Figure \ref{fig:chan vese inputs} and \ref{fig:chan vese inputs 2} we show the results with a superimposed mask of the computed region $D_1$ for some details cropped from the original images. 

Because of the intrinsic limitations of the Chan-Vese approach, we observe that the segmentation result is not satisfactory (see, for instance, Figure \ref{fig:chan vese inputs}) since it generally detects with high precision only the largest uniform region around the user selection. To detect the whole inpainting domain $D$, the user should give in principle many initialisation points, which may be very demanding in the presence of several disconnected and possibly tiny inpainting regions.

For this reason, we proceed differently and make use of a feature-based approach to 
use the area $D_1$ as a training region for a classification algorithm running over the whole set of image pixels.

\subsubsection{Feature extraction.} \label{sec:features}
We now extract some intensity-type features from the image in order to build feature vectors which will be then used for classification. 
Namely, for every pixel $x$ in the image we apply non-linear channel transformations to compute the HSV (Hue, Saturation, Value), the geometric mean chromaticity GMCR \cite{Finlayson2009}, the CIELAB and the CMYK (Cyan, Magenta, Yellow, Key) values (see \cite{wyszecki2000color} for more details). 
Once this is done, we append all these values together so that multi-dimensional feature vector $\bm{\psi}$ of the form
\begin{equation}  \label{eq:features}
\bm{\psi}(x)= [ \textrm{HSV}(x), \textrm{GMCR}, \textrm{CIELAB}(x), \textrm{CMYK}(x)]
\end{equation}
is built. 
Note that having only the RGB intensity data at hand, unfortunately not much information can be added when building the feature vectors. 
In this respect, if one would have multi-spectral measurements (such as IR, UV or other types) additional features extracted from the complementary data could be added for an improved result.

\subsubsection{A classification algorithm with training}  \label{sec:k-means}
Once the feature vectors are built for every point in the image, we use the training region $D_1$ detected as described in Section \ref{sec:chan-vese} as a dictionary to drive the segmentation procedure extended to the whole image domain. 
We proceed as follows. 
First, we run a labelling algorithm over the whole image domain comparing appropriately the features defined in \eqref{eq:features} in order to partition the image in a fixed number of $K$ classes. 
To do that, we use the well-known $k$-means algorithm. 
After this preliminary step, we check which label has been assigned to the training region $D_1$ and simply identify in the labelled image which pixels share the same label. 
By construction, this corresponds to find the regions in the image `best-fitting' the training region in terms of the features defined in Section \ref{sec:features}, which is our objective. 
After a refinement step for the removal of possibly small details due to noise and/or misclassification, we can finally extract the whole area to inpaint $D$. 
We report the results corresponding to Figure \ref{fig:chan vese inputs} and Figure \ref{fig:chan vese inputs 2} in Figure \ref{fig:k means segm} and \ref{fig:k means segm 2}, respectively.

Note that in order to reduce the computational times, our results are run not on the whole image, but only on details cropped by the user as a very first step. 
Additionally, in order to provide a suitable initialisation for the Chan-Vese segmentation, the user (typically, an expert) is required to select few `training' pixels in the inpainting domain to guide the first segmentation procedure. 
As such, the method is semi-supervised.

\subsection{Inpainting models}   \label{sec:exemplar inpainting}
Once an accurate segmentation of the damaged areas is provided, the task becomes the actual restoration of the image contents in $D$ by means of the available information in the region $\Omega\setminus D$.  
The standard mathematical approach solving an inpainting problem consists in minimising an appropriate function $\mathcal{E}$ defined over the image domain $\Omega$, i.e.\ in solving 
\begin{equation} \label{eq:minimisation}
\text{ find }\qquad u\qquad\text{s.t.}\qquad u\in \text{argmin}_v~ \mathcal{E}(v).
\end{equation}
A standard choice for $\mathcal{E}$  in the case of \emph{local} inpainting models is the functional 
\begin{equation}  \label{eq:functional}
\mathcal{E}(v) = R(v) + \lambda\chi_{\Omega\setminus D}\| f- v\|^2_2,
\end{equation}
where $f$ denotes the given image to restore, $\|\cdot\|_2$ is the Euclidean norm and $\chi_{\Omega\setminus D}$ denotes the characteristic function of the unoccluded image areas, so that for every pixel $x\in \Omega$:
$$
\chi_{\Omega\setminus D}(x) = \begin{cases}
1\quad&\text{if }\quad x\in \Omega\setminus D\\
0 \quad & \text{if }\quad x\in D.
\end{cases}
$$
Because of this modelling, the second term in \eqref{eq:functional} can be interpreted as a distance function between the given image $f$ in the intact and the unknown image to reconstruct. 
Indeed, the multiplication by the characteristic function $\chi$ implies that such term is simply zero for the points in $D$, since there no information is available, while the contribution of all the points in $\Omega\setminus D$ has to be as small as possible. 
The term $R$ typically encodes local information (such as gradient magnitude) which is the responsible of the transfer of information inside $D$ by means of possibly non-linear models\cite{ChanShenBook,CarolaBook}. 
The transfer process is balanced with the trust in the data by the positive parameter $\lambda$. 
A classical choice of a gradient-based inpainting model consists in choosing
\begin{equation}  \label{eq:TV}
R(v) = \| \nabla v \|_1 = \sum_{x\in\Omega} | \nabla v(x) | 
\end{equation}
i.e.\ the Total Variation of $v$ \cite{ShenChan2002}. As mentioned above such an image inpainting technique is not designed to transfer texture information. Furthermore, it fails in the inpainting of large missing areas. For our purposes we use \eqref{eq:TV} as an initial `good' guess with which we initialise a different approach based on a local copy-and-paste procedure as described in the following section.

\subsubsection{Exemplar-based inpainting.} \label{sec:nonlocal inpainting}
We describe here the non-local patch-based inpainting procedure studied in \cite{Arias2011,Newson2014} and carefully described in \cite{NewsonIPOL} from an implementation point of view
\footnote{The code is freely available at IPOL: \href{https://doi.org/10.5201/ipol.2017.189}{https://doi.org/10.5201/ipol.2017.189}}.
In the following, we define for any point $x\in \Omega$  the \emph{patch neighbourhood} $\mathcal{N}_x$ as the set of points in $\Omega$ in a neighbourhood of $x$. 
Assuming that the patch neighbourhood has cardinality $n$, by \emph{patch} around $x$ we denote the $3n$-dimensional vector $P_{x} = (u(x_1), u(x_2),\ldots,u(x_n) )$  where the points $x_i, i=1,\ldots n$ belong to patch neighbourhood $\mathcal{N}_x$. 
In order to measure `distance' between patches, a suitable patch measure $d$ can be defined, so that $d(P_{x},P_{y})$ stands for the patch measure between the patches around the two points $x$ and $y$. We define then the Nearest Neighbour (NN) of $P_{x}$  as the patch $P_y$ around some point $y$ minimising $d$. 

For an inpainting application the task consists then in finding for each point $x$ in the inpainting domain $D$ the best-matching patch $P_y$ outside $D$.
Assuming that each NN patch can be characterised in terms of a shift vector $\phi$ defined for every point in $\Omega$ (i.e.\ assuming there exists a  rigid transformation $\phi$ which shifts any patch to its NN), the problem can be formulated as the minimisation problem
\begin{equation}   \label{eq:nonlocal inp}
\min ~ \mathcal{E}(u,\phi) = \sum_{x\in D}~ d^2(P_{x},P_{x+\phi(x)}).
\end{equation}
Heuristically, every patch in the solution of the problem above is constructed in such a way that in the damaged region $D$ has a correspondence (in the sense of the measure $d$) with its NN patch in the intact region $\Omega\setminus D$. Following \cite{NewsonIPOL}, we use the following distance:
\begin{equation}   \label{eq:measure}
d^2(P_{x},P_{x+\phi(x)})= \sum_{y\in \mathcal{N}_{x}} (u(y)- u(y+\phi(x)))^2.
\end{equation}
From an algorithmic point of view, solving the model above corresponds to consider mainly two phases: the former consists in computing (approximately) the NN patch for each point in $D$, so as to provide a complete representation of the shift map $\phi$ This can be computationally expensive for large image data. 
In order to solve this efficiently, a PatchMatch \cite{Barnes2009} strategy can be applied. 
Afterwards a proper image reconstruction step is performed, where for every point in $D$ the actual corresponding patch is computed. 
We refer the reader to \cite{NewsonIPOL} for the numerical details. 

A crucial ingredient for a good performance of the non-local algorithm is its initialisation. 
In particular, once the inpainting domain is known, a pre-processing step where a local inpainting model, such as the TV inpainting model \eqref{eq:functional} with \eqref{eq:TV}, can be run to provide a rough, but reliable initialisation of the algorithm\footnote{The code is freely available at IPOL: \href{https://doi.org/10.5201/ipol.2012.g-tvi}{https://doi.org/10.5201/ipol.2012.g-tvi}}. 

We report the results of the combined procedure in Figure \ref{fig:inpainting} and the overall work-flow of the combined algorithm in the diagram  in Figure \ref{fig:workflow segmentation}.

\subsubsection{Model parameters} 
For the segmentation of the the training region $D_1$ within the inpainting domain $D$ we use the \texttt{activecontour} MATLAB function by which the Chan-Vese algorithm can be called. 
For this we fixed the maximum number of iterations to \texttt{maxiter}$=1000$ and use the default value as a tolerance on the relative error between iterates as a stopping criterion. 
We use the default values for the parameters $\mu$ and $\nu$ in \eqref{eq:chanvese}. 
The subsequent labelling phase was performed by means of the standard MATLAB \texttt{kmeans} function after specifying a total of $K=35$ labels to assign.
The labelling was iteratively repeated $5$ times to avoid misclassification. Once the detection of the inpainting domain is completed, in order to provide a good initialisation to the exemplar-based model we use the TV inpainting model \eqref{eq:minimisation} with \eqref{eq:TV} with the value $\lambda=1000$ and a maximum number of iterations equal to \texttt{maxiter2}$=1000$ with a stopping criterion on the relative error between iterates depending on a default tolerance. 
Finally, we followed \cite{NewsonIPOL} for the implementation of the  exemplar-based inpainting model: for this we specified $12$ propagation of iterations and tested different size of the patches. In order to avoid memory shortage, we restricted to patches of the size $5\times 5$, $7\times 7$ and $9\times 9$. 

The numerical tests were performed here on a standard MacBook Pro (Retina, 13-inch, Early 2015), 2,9 GHz Intel Core i5, 8 GB 1867 MHz DDR3 using MATLAB 2016b.

\subsection{Discussion and outlook} \label{sec:inpaintingdisc}
We proposed in this section a combined algorithm to retrieve image contents from two images of illuminated manuscripts shown in Figure \ref{fig:illuminated manuscripts} where very large regions have been damaged.
At first, our algorithm computes an accurate segmentation of the inpainting domain which is performed by means of a semi-supervised method exploiting distinctive features in the image. 
Then, taking the segmentation result as a input, the procedure is followed by an exemplar-based inpainting strategy (upon suitable initialisation) by which the damaged regions are filled by using sensible patches. 

The results reported in Figure \ref{fig:inpainting} and \ref{fig:inpainting 2} confirm the effectiveness of the combined method proposed. 
In particular, when looking at the difference between standard local (TV) image inpainting methods and the non-local one, we immediately appreciate the higher reconstruction quality in the damaged regions, especially in terms of texture information. 
The method has been validated on several image details extracted from the entire images, and has been shown effective also for very large image portions with highly damaged regions.

In term of computational times, the segmentations in Figure \ref{fig:k means results} are obtained in approximatively 15 minutes. The inpainting results in Figure \ref{fig:inpainting} are obtained in about 3 minutes for patches of size 5x5 and about 7 minutes for patches of size 7x7. 
Overall the whole task of segmenting and inpainting the occluded regions takes approximatively 20 minutes per images of size 690x960, approximatively.
However, these results highly depends on the size of the image, the
size of the inpainting domain and the size of the patches chosen.

Future work could address the use of different features for the segmentation of the inpainting domain with similar methodologies, such as for instance texture features \cite{Varma2005}. 
Furthermore, at an inpainting level, we observe that the reconstruction of fine details in very large damaged regions (such as the strings of the harp in Figure \ref{fig:inpainting 2}) is very challenging due to the lack of correspondence with similar training patches in the undamaged region. 
For solving this problem a combination of exemplar-based and structure-preserving inpainting models could be used.

\section{Looking through the layers via osmosis filtering}
\label{sec:osmosis}
In the previous section, the content of the damaged areas is completely lost
and can only be estimated by assuming a content similar to the rest of the
picture.  But this is not the only kind of degradation encountered in
illuminated manuscripts.  In some cases parts of an illumination are painted
over. The illuminations displayed in this section are from the primer of Claude
de France. A prudish later owner could not stand Adam and Eve being nude and
had the offensive areas painted over. In cases like this one, we can use
infrared information to actually look through the layers and recover the
apparently lost information.  In the following we present the linear osmosis
model, how it is generally used and  we apply it to the illuminations from the
primer.

\subsection{The Osmosis model}
\label{subsec:osmosismodel}
The osmosis model
has been introduced in \cite{weickert2013linear} as a non-symmetric
generalization of diffusion filters and a new tool for image processing
problems such as seamless cloning and shadow removal. The original parabolic
equation for this model is
\begin{equation} 
    \label{eq:parabolicosmosis} 
    u_{t} = \Delta u - \mathrm{div}(\mathbf{d}u) 
\end{equation}
where $\mathbf{d}$ is the input vector field defined on the image domain
$\Omega$ with values in $\mathds{R}^{2}$.  The equation is typically solved on
the whole image with homogeneous Neumann boundary conditions and this model has
already been successfully used to solve, by dimensional splitting techniques, the light balance problem in
Thermal-Quasi Reflectography (TQR) imaging, a non-destructive way to support
the restoration of large mural paintings \cite{DAFFARA2018266,parisotto2018digital}.

For the work presented in the following however, we look  directly for the
solution of the stable state of the previous equation, the elliptic equation,
\begin{equation} 
    \label{eq:ellipticosmosis} 
    \Delta u = \mathrm{div} (\mathbf{d}u) 
\end{equation}
on a small sub-domain $D$ of the input image $\Omega$,
with mixed boundary conditions. Restricting ourselves to a small domain has
two main advantages: first most of the image is supposed to already be the
ground truth and is therefore best left untouched  and second the
computation speed is much faster. Moreover having mixed boundary conditions
allows for more flexibility depending on the problem at hand. On the one
hand Dirichlet boundary conditions enforce the colour values on $\partial D$ and
a smooth transition when there shouldn't be any discontinuity with the
untouched part of the image.  On the other hand Neumann boundary conditions
prevent any diffusion across the boundary, ensuring clear colour
discontinuities which is useful when the border of the mask is along an
edge between two different colours.

The drift vector field $\mathbf{d}$ encodes information from the gradient of
the desired solution and serves as a guide to the diffusion. When $\mathbf{d} =
\nabla I/I$, $I$ is an evident solution of the equation and this vector field
is then called the canonical drift vector field of the image $I$.  In the
following, we will note $\mathbf{d}_{I}$ the canonical drift vector field of
the image $I$. The final method used for the primer is a combination of the two following applications.

\subsubsection{Application to seamless cloning}
\label{subsec:seamlesscloning}
Different results come from different manipulations of this canonical vector
field.  The problem of seamless cloning can be described as an improvement of a
copy-and-paste process where the pasted part is modified in such a way that the
pasting is indistinguishable.  It typically involves two images, the background
one and the one from which comes the pasted region.  To seamlessly paste some
area of an image $f$ on the domain $D$ of an image $g$, the drift vector field
is created by replacing $\mathbf{d}_{g}$ by $\mathbf{d}_{f}$ on the domain $D$
and taking the mean of both on the boundary $\partial D$ and the equation is
then solved either on the whole image domain $\Omega$ with Neumann boundary
conditions on $\partial\Omega$ or only on the sub-domain $D$ with Dirichlet
boundary conditions.

\subsubsection{Application to shadow removal}
\label{subsec:shadowremoval}
The
problem of shadow removal involves only one image and is, as its name
indicates, a process that takes an image with shadowed areas and gives as a
result the same view without any shadows.  A shadow can be reduced as a
multiplicative change in the domain of the shadowed region of the image while
the canonical drift vector field is invariant to multiplicative change. The
presence of the shadow is therefore only encoded in the drift vector field on
the edge of the shadow.  In an ideal case with a sharp constant shadow,
negating the drift field on the edge of the shadow creates pure diffusion on
the edge and results in a perfect shadowless image.

\subsection{Application to the illuminations}
\label{subsec:osmosisillumination}
In the ideal case, the added pigments do not
appear on the infrared image while the colours to be restored are perfectly
encoded in the infrared image. In this case the problem is reduced to a simple
seamless cloning application with Dirichlet boundary conditions. The
drift-field of the RGB (Red Green Blue) image is replaced by the one from the infrared image on the
sub-domain to be restored.  But such an ideal case is  uncommon. For the
illuminations of the primer, several issues with the infrared appear. In most
cases, the osmosis model is enough to get a nice output. For more challenging
cases, we have to add some workaround.

\subsubsection{Using only osmosis}
A problem encountered in Figure
\ref{fig:adameveleft} is that green pigments are not well encoded in the
infrared image. This has the nice effect that the green leaves added are
completely absent from the infrared image. Yet the original fig leaf is
also green and the colour distinction between the fig leaf and the skin is
slight on the infrared image. Therefore simply following the seamless cloning
method doesn't lead to satisfactory results. On the RGB image, the
original fig leaf colour is present. To enforce that the skin colour and the
green of the original fig leaf do not mix, we have to add Neumann boundary
conditions along the edges involved. This is illustrated in Figure
\ref{fig:adameveleft}, the Neumann boundary conditions being enforced along the
red lines in the mask.

Another issue encountered is that the added pigments often appear on the
infrared image.  If the added cloth has little to no texture discernible on the
infrared and the original image appears by transparency, a mix of
seamless cloning and shadow removal is applied. The added cloth is treated as a shadow.
The drift-field of the RGB image is replaced by the drift-field of the
infrared image on the sub-domain to be restored but the drift-field is put to zero
on the edge of this sub-domain. This method is illustrated in Figure
\ref{fig:adameveright}. The white lines of the mask are the areas where the
drift-field is put to zero.  If a texture from the added cloth is present in the
 infrared image, it appears in the final result. In Figure \ref{fig:adameveright},
the result is still visually acceptable.

\subsubsection{Adding an inpainting step}
In the case of Figure
\ref{fig:adamleft}, the infrared image adds some useful information to the RGB image.
Yet, a large amount of the added skirt texture appears on the
infrared. This prevents us from using the process previously described.  To get some
results, we  manually draw the main edges appearing in the infrared and use
this hand drawn sketch as the second image for the seamless cloning process.
However, this leads to a complete loss of texture in the inpainted domain.
Therefore to get back texture in the restored area, we use this intermediate
result as initialisation for the exemplar-based inpainting described in the
previous section. This gives more natural   outputs but the results are
not as satisfying as the ones displayed in Figure \ref{fig:adameveleft} and
\ref{fig:adameveright}.

\subsection{Preprocessing and parameters}
As we  just saw, such a complex restoration process necessitates significant user decisions. In fact the mask containing the sub-domain to
be restored must be drawn by the user as well as the edges along which Neumann boundary conditions should be
applied and the sub-domain edges where the drift-field should be put to zero.

For our experiments we used the discretisation proposed in
\cite{weickert2013linear}. Then the linear system was solved using the MATLAB
UMFPACK V5.4.0 LU solver. It took  8.61 seconds to numerically obtain the
final result of Figure \ref{fig:adameveright} and 6.10 seconds for the final
result of Figure \ref{fig:adameveleft}.  For the inpainting step of Figure
\ref{fig:adamleft}, we used the implementation of the exemplar-based inpainting
algorithm from \cite{ipol.2015.136}
\footnote{The code is freely available at
IPOL:
\href{https://doi.org/10.5201/ipol.2015.136}{https://doi.org/10.5201/ipol.2015.136}}
with the nlmedians method, $9\times9$ patches, two scales and 3 iterations.

The numerical tests were performed on a standard MacBook Pro (Retina, 13-inch,
2017), 3,5 GHz Intel Core i7, 16 GB 2133 MHz LPDDR3 using MATLAB 2017b.

\subsection{Discussion and future work} \label{sec:osmosisdisc} We proposed in
this section a method to restore covered up parts of a manuscript using 
infrared information. For the illuminations from the primer that we worked on,
the results are mostly satisfying, especially when the added pigments do not
appear on the infrared or when the addition doesn't have too much texture. 
This method necessitates some patient work for the mask creation
and is heavily dependent on the infrared wavelength. Moreover it hasn't been
tested on other illuminated manuscripts.

Future work should address these difficulties and test the method on a larger
dataset. For the mask creation phase, a semi-supervised segmentation step could be inserted. In this work, we have only used the visible image and a single infrared image. Some other multispectral images may help in distinguishing the added texture from the original one. But the expert would have to specify for each area which images should be used.

\section{Creating a 3D virtual scene from illuminated manuscripts}  \label{sec:3D visualisation}

Having restored a damaged manuscript based on the techniques in the proceeding sections, we can now go a step further and build a 3D reconstruction.  Specifically, we convert the manuscript Simon Bening, \emph{Annunciation}, Fitzwilliam Museum, MS 294b, Flanders, Bruges, (1522-1523) (shown in Figure \ref{fig:twoEyes}(a)) into stereo 3D.  We do so using a 3D conversion pipeline originally developed for the conversion of Hollywood films.  There, one is given the video shot from camera position $p \in \field{R}^3$ and orientation $O \in SO(3)$ (corresponding to, for example, the left eye view), and the objective is to generate a plausible reconstruction of the the video as it would appear from a perturbed position and orientation $p+\delta p \in \field{R}^3$, $O+\delta O \in SO(3)$ (corresponding to the view from the other eye).  In some cases $p$ and $O$, along with other relevant camera parameters such as field of view, may be given.  In other cases, they must be estimated.  In our case the process is the same, except that we have a manuscript rather than a video.  However, this introduces is a subtle difference. Before, although we might not know the camera parameters associated with a given video, we at least knew that they {\em exist} - but here, because the input is drawn by a human, we cannot take existence for granted.  In particular, depending on the artist, the drawing may or may not obey the laws of perspective.  We will show an example of this at work in the famous painting {\em The Scream} by Edvard Munch in Figure \ref{fig:scream}.  

\subsection{Overview of a 3D Conversion pipeline} 

\label{sec:pipeline}

Here we briefly go over the 3D conversion pipeline used in this paper.  For more details, please see \cite{Guidefill} or \cite[Ch. 9.4]{CarolaBook}.  Note that in industry, this pipeline is executed by teams of artists each responsible for different stages.  Each stage is either done with a toolkit of algorithms that are then touched up by hand, or is done completely manually.  The basic workflow is as follows:
\begin{enumerate}
\item Generate a rough but plausible 3D model of the scene, including a virtual camera with plausible parameters (parameters include position, orientation, field of view, possibly lense distortion, etc) placed within it.  In industry, artists may be given a 3D model of the scene directly from the client, or may build it themselves.  For scenes with motion, camera parameters can be estimated with the matchmove algorithm \cite{MatchMove,MatchMove2}.  The 3D models do not have to be perfect, and are typically made to a little larger than the objects they correspond to.  This is because they will be ``clipped'' in the next step.
\item Generate accurate masks for all objects in the scene.  This is typically done by hand, but could also be done with the help of segmentation algorithms that are then touched up.  These masks are then used, much like a cookie cutter, to ``clip'' the 3D geometry (which, remember, was generated to be a little bit too large), throwing away the part of the geometry that is unneeded (an illustration is provided in Figure \ref{fig:geometry}).
\item The camera is transformed into a projector, projecting the input image (or video) onto the 3D geometry.
\item One or more new virtual cameras are added to the scene.  If the original camera is taken to be either the right or left eye, then one additional virtual camera corresponded to the other eye is needed.  However, sometimes the original camera position is taken to be half way between the two eyes, so that two virtual cameras (corresponding to the left and right eyes) are needed.  These camera(s) will be used to render the 3D scene from one or more new viewpoints, in order to create a stereo pair.
\item Because the new camera(s) will typically see bits of background previously hidden behind foreground objects in the original view, inpainting of occluded areas is required.  This is typically done using a toolbox of inpainting algorithms that are then touched up by hand.  In our example, inpainting was done in Photoshop, using a combination of Content Aware fill and manual copy pasting of patches by hand.
\end{enumerate}
Steps one and two can be thought of as generating a {\em depth map} for the image.  The rough geometry generated in step one provides the smooth component of the depth map, while the masks generated in step two generate the depth discontinuities.  Because the human eye is most sensitive to depth discontinuities, these have to be very accurate, but the 3D models do not.  For example, in the conversion of Figure \ref{fig:twoEyes}(a), the virgin Mary is modelled using just a few simple geometric primitives including an ellipsoid for her body, a sphere for her head, a cylindrical halo and a cone for the bottom of her dress.  This is illustrated in \ref{fig:geometry}, where we also show the geometry for the angel Gabriel.

\subsection{Results and future work}\label{sec:3Ddisc}

The results of our 3D conversion are presented in Figure \ref{fig:twoEyes}, where we show the original manuscript (assumed to be the right eye view) side by side with the reconstructed left eye view. A limitation of the pipeline we have used is that it does not handle partially transparent objects properly.  In this case, bits of background in the original right eye view are visible through the halos of both the virgin Mary and the angel Gabriel.  In particular, in the original right eye view, a bit of Mary's bed is visible through her halo.  When rendered from the new left eye vantage point, we should now be seeing the window through her halo, but instead we continue to see the bed.  See Figure \ref{fig:limitation} for a closeup of this defect. To overcome this, one could modify the pipeline in Section \ref{sec:pipeline} to first decompose semi-transparent objects into two images (in this case, the pure halo and the background).  This is something we would like to investigate in the future. 

\section{Conclusion}
An adequate mathematical analysis and processing of images arising in the arts and humanities needs to meet special requirements:
\begin{itemize}
\item There is often particular domain expertise which any analysis should ideally make use of. For instance, when digitally restoring an image, the integration of related images such as paintings from the same artist, could be taken into account. In what we have discussed this concept is used to the extent that a dictionary of characteristic structures in the undamaged part of the illuminations was created and used to fill in the lost contents in the damaged regions, compare Figures \ref{fig:inpainting}, \ref{fig:inpainting 2}. This could be driven much further, expanding the dictionary by illuminations or details of illuminations from the same artist. 
\item The results achieved in Figures \ref{fig:adameveleft}, \ref{fig:adameveright}, \ref{fig:adamleft} show a possible use-case for scientific imaging in art restoration or art interpretation. Indeed, we believe that the integration of different types of scientific imaging such as infrared imaging, are likely to give benefit to image analysis methods and so the latter should be able to capture those.
\item Explainability of results is crucial. There is clearly a balancing act to be made between hand-crafted analysis that captures expert knowledge and a black-box, data-driven image analysis approach. In particular, the latter should ideally have an interpretable mathematical representation that gives rise to new conclusions. In this paper we have solely considered model-based and hence explainable solutions to art restoration and interpretation problems. The growing emergence of deep learning solutions to various image analysis tasks provides an alternative approach to these problems, at the moment however without a proper explanation.
\item Relevant characteristics are often hidden in very fine details of the artwork, like a brushstroke in a painting. Capturing these fine details in a digital format results in high-resolution images that an image analysis method should be capable of processing. This means there is a demand for computationally-efficient image analysis methods. 
\end{itemize}
With the above in mind, we have discussed a selected subset of mathematical approaches and their possible use-cases in the restoration and interpretation of illuminated manuscripts. These approaches are not perfect yet by all means and there is plenty of room for improvement, compare our discussion in Sections \ref{sec:inpaintingdisc}, \ref{sec:osmosisdisc} and \ref{sec:3Ddisc}. 

\section*{Acknowledgements}
Luca Calatroni acknowledges the support of Fondation Math\'emathique Jacques Hadamard (FMJH). 
Simone Parisotto acknowledges the UK Engineering and Physical Sciences Research Council (EPSRC) grant EP/L016516/1 for the University of Cambridge Centre for Doctoral Training, the Cambridge Centre for Analysis. Carola-Bibiane Sch\"oenlieb acknowledges support from the Engineering and Physical Sciences Research Council (EPSRC) 'EP/K009745/1', the Leverhulme Trust project 'Breaking the non-convexity barrier', the EPSRC grant 'EP/M00483X/1', the EPSRC centre 'EP/N014588/1', the Alan Turing Institute 'TU/B/000071', CHiPS (Horizon 2020 RISE project grant), the Isaac Newton Institute, and the Cantab Capital Institute for the Mathematics of Information  


\begin{backmatter}

\section*{Competing interests}
The authors declare that they have no competing interests.

\section*{Author's contributions}
All authors contributed to the manuscript in equal parts.


\bibliographystyle{bmc-mathphys} 
\bibliography{biblio}      




\section*{Figures}

  \begin{figure}[h!]  
  \includegraphics[width=5cm,trim=0cm 0cm 5.7cm 0cm,clip=true]{\detokenize{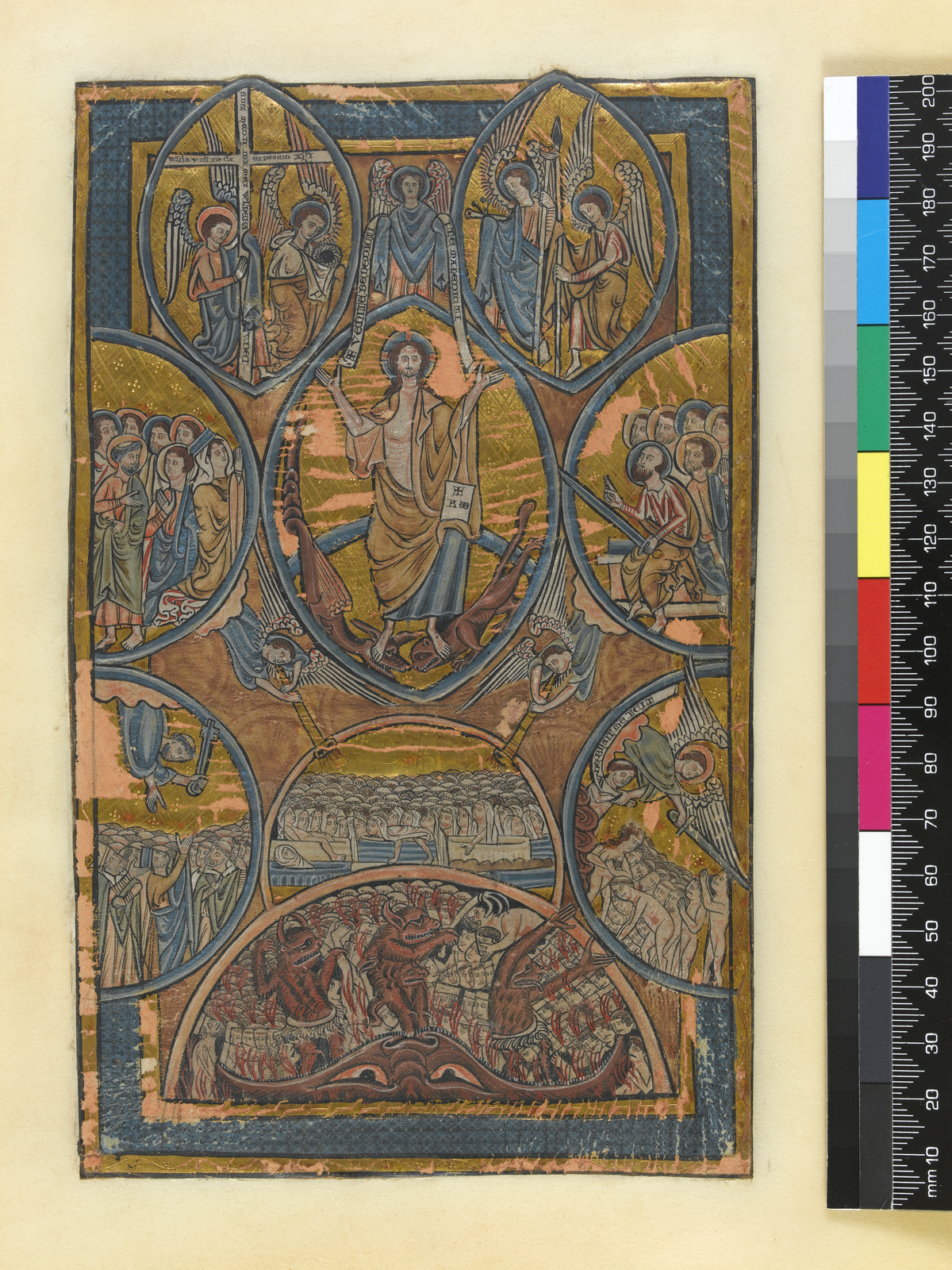}} \qquad 
  \includegraphics[width=5cm,trim=0cm 0cm 5.7cm 0cm,clip=true]{\detokenize{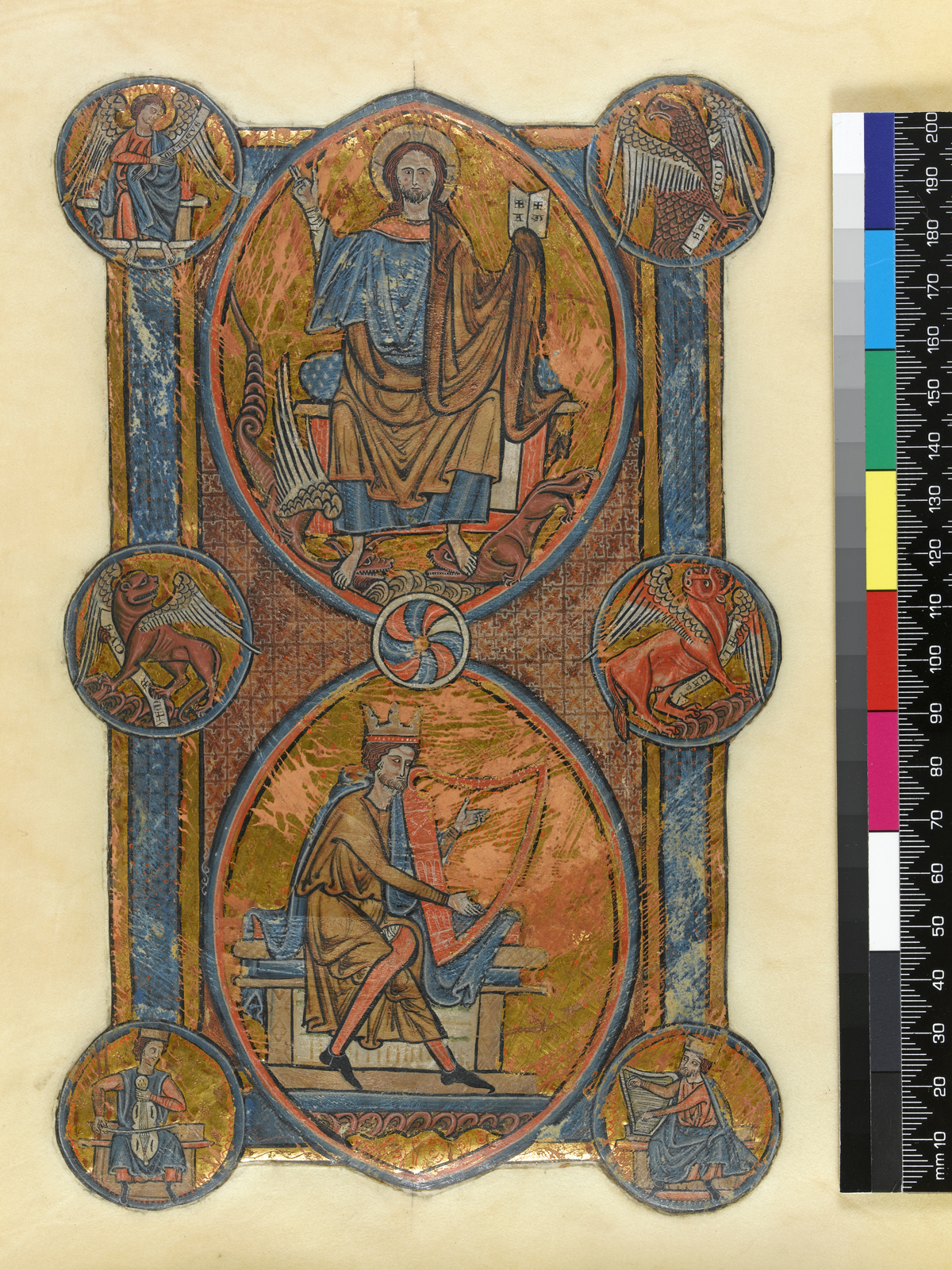}} 
  \caption{\csentence{Illuminated manuscripts.} William de Brailes, Last Judgement (left) and Christ in Majesty with King David playing the harp (right). Fitzwilliam Museum, MSS 330.iii and 330.v.\ England, Oxford, c.\ 1230-1250.
      These two illuminated manuscripts show large and non-homogeneous damaged areas, see Section \ref{sec:illuminated_manuscripts} for more details.}
      \label{fig:illuminated manuscripts}
      \end{figure}
      
       \begin{figure}[h!]  
  \begin{subfigure}{0.31\textwidth}
\begin{center}
\includegraphics[height=4cm]{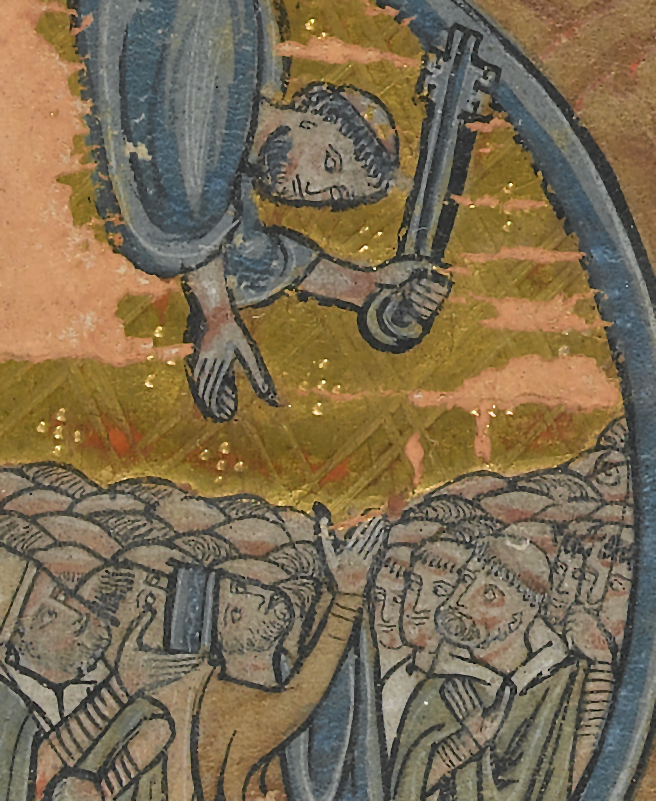}
\end{center}
\caption{Detail}
  \end{subfigure}
    \begin{subfigure}{0.31\textwidth}
\begin{center}
\includegraphics[height=4cm]{\detokenize{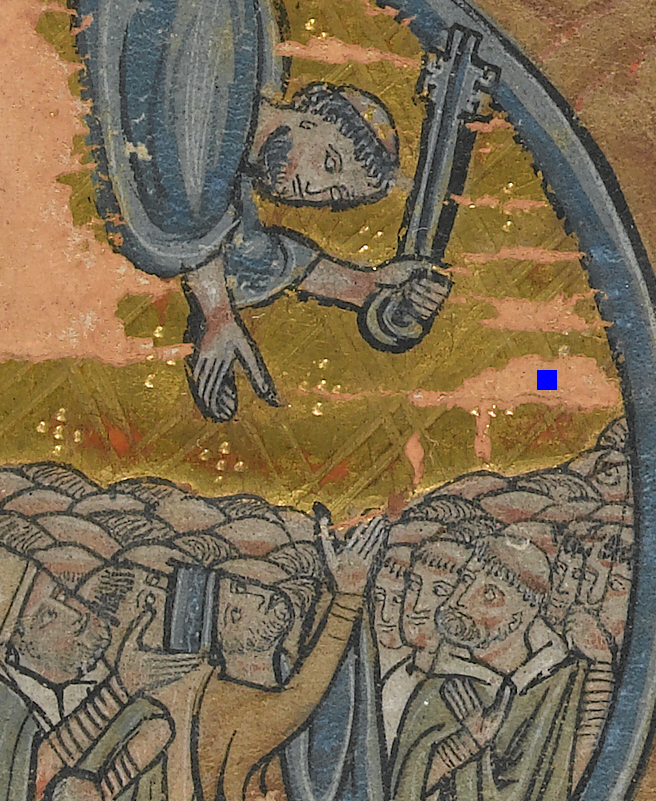}}
\end{center}
\caption{User selection}
  \end{subfigure}
      \begin{subfigure}{0.31\textwidth}
\begin{center}
\includegraphics[height=4cm]{\detokenize{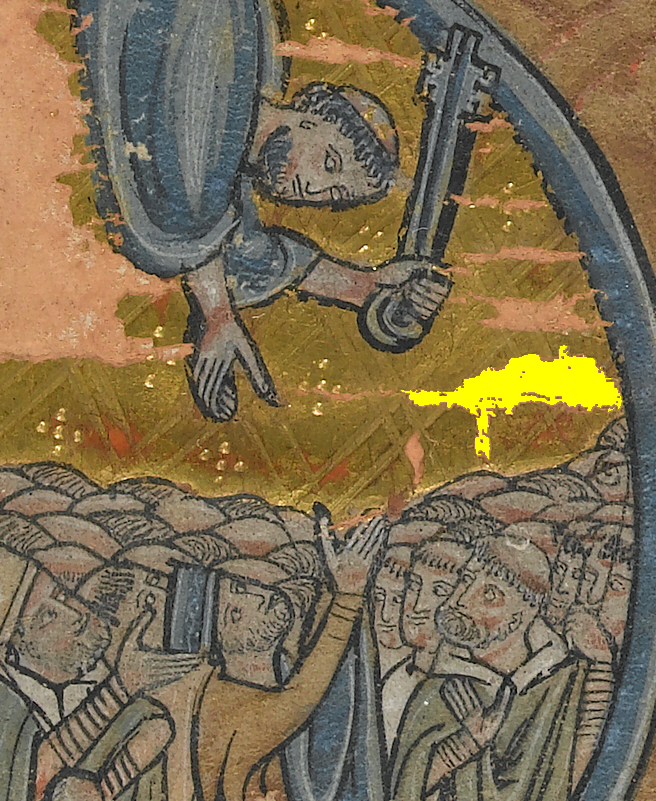}}
\end{center}
\caption{Chan-Vese segmentation}
  \end{subfigure}
  \caption{\csentence{Detection of the training region.}
     The user clicks on the damaged region to select training pixels (in blue) which serve as initialisation of the Chan-Vese model \eqref{modelchanvese}. The segmentation algorithm is run and a training region inside the damaged area is segmented accurately. The result is superimposed to the initial image and coloured yellow for better visualisation.}
      \label{fig:chan vese inputs}
      \end{figure}
      
             \begin{figure}[h!]  
  \begin{subfigure}{0.31\textwidth}
\begin{center}
\includegraphics[height=4cm]{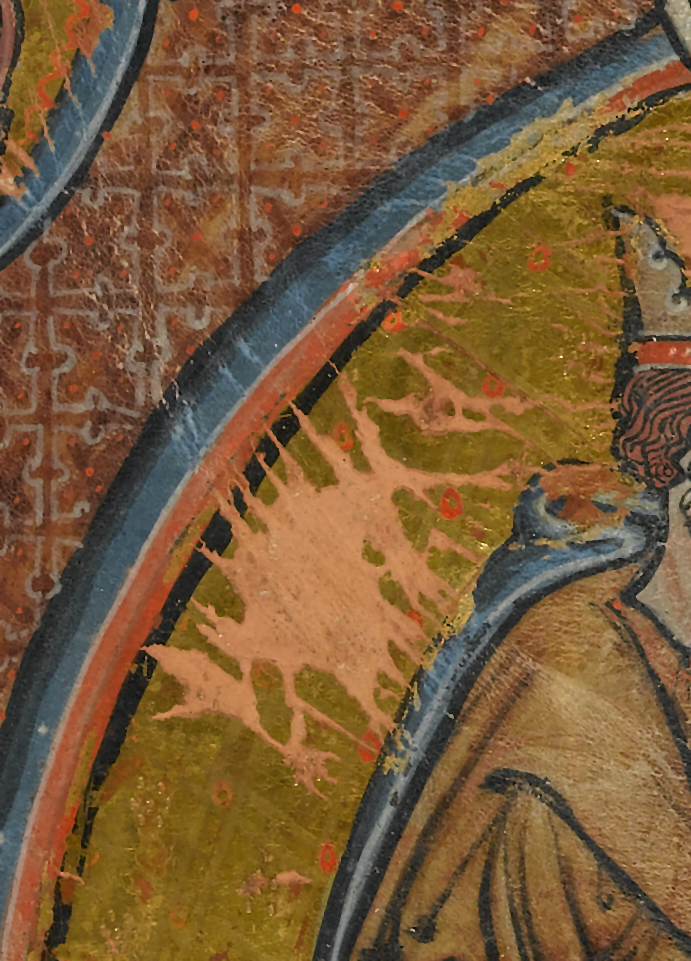}
\end{center}
\caption{Detail}
  \end{subfigure}
    \begin{subfigure}{0.31\textwidth}
\begin{center}
\includegraphics[height=4cm]{\detokenize{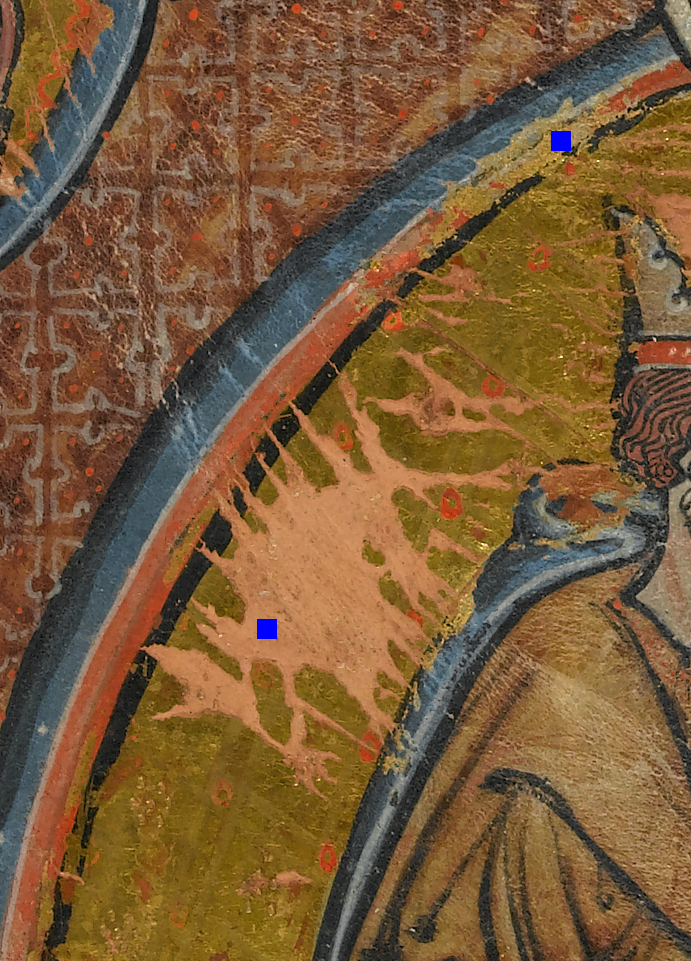}}
\end{center}
\caption{User selection}
  \end{subfigure}
      \begin{subfigure}{0.31\textwidth}
\begin{center}
\includegraphics[height=4cm]{\detokenize{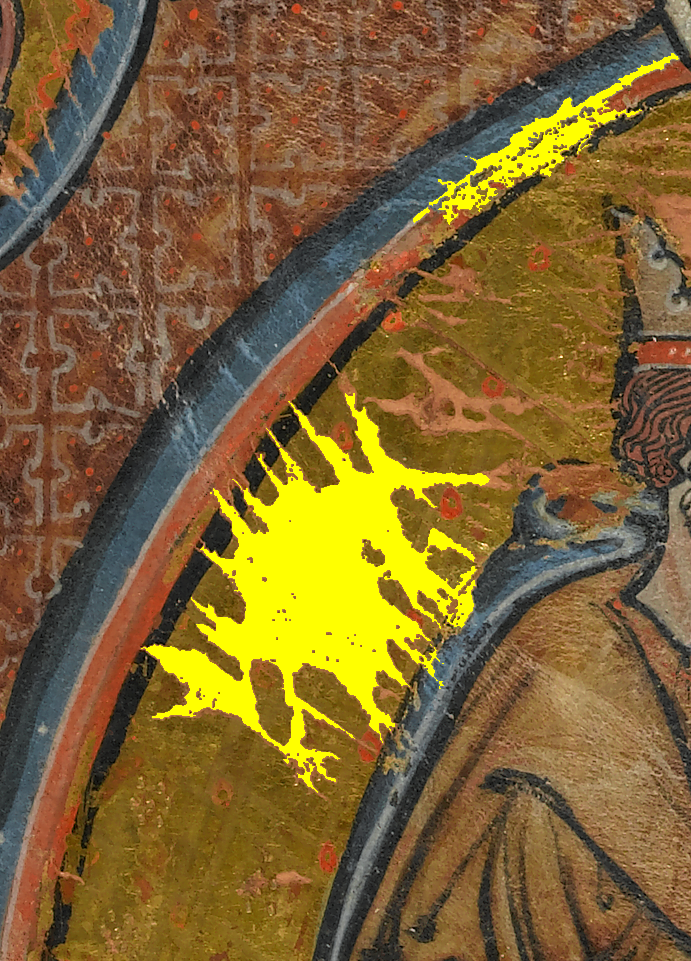}}
\end{center}
\caption{Chan-Vese segmentation}
  \end{subfigure}
  \caption{\csentence{Detection of multiple training regions.}
     The user clicks on each damaged region to select training pixels (in blue) which serve as initialisation of the Chan-Vese model \eqref{modelchanvese}. The segmentation algorithm is run and multiple training regions are selected. The result is superimposed to the initial image and coloured yellow for better visualisation.}
      \label{fig:chan vese inputs 2}
      \end{figure}

           \begin{figure}[h!]  
  \begin{subfigure}{0.45\textwidth}
\begin{center}
\includegraphics[height=4cm]{\detokenize{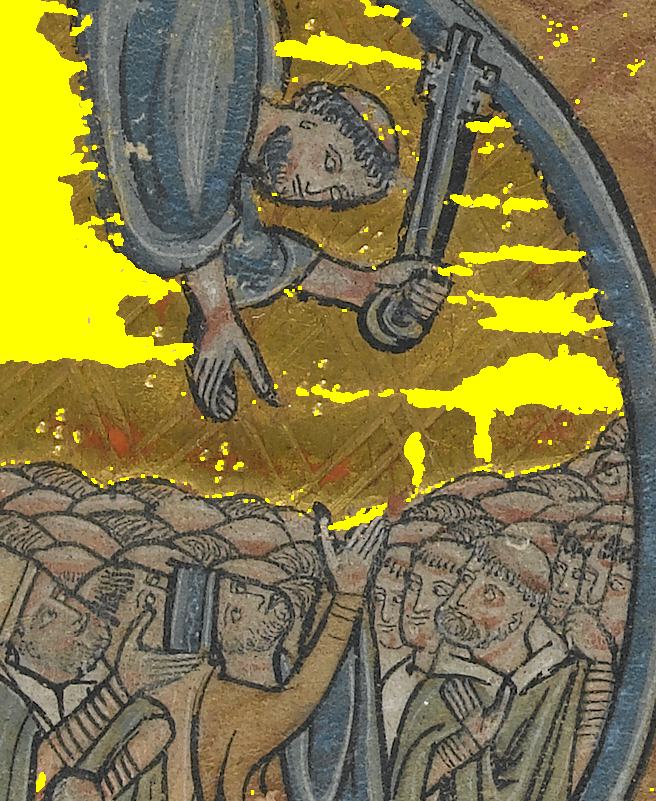}}
\end{center}
\caption{Detail from Figure \ref{fig:chan vese inputs}.}
\label{fig:k means segm}
  \end{subfigure}
    \begin{subfigure}{0.45\textwidth}
\begin{center}
\includegraphics[height=4cm]{\detokenize{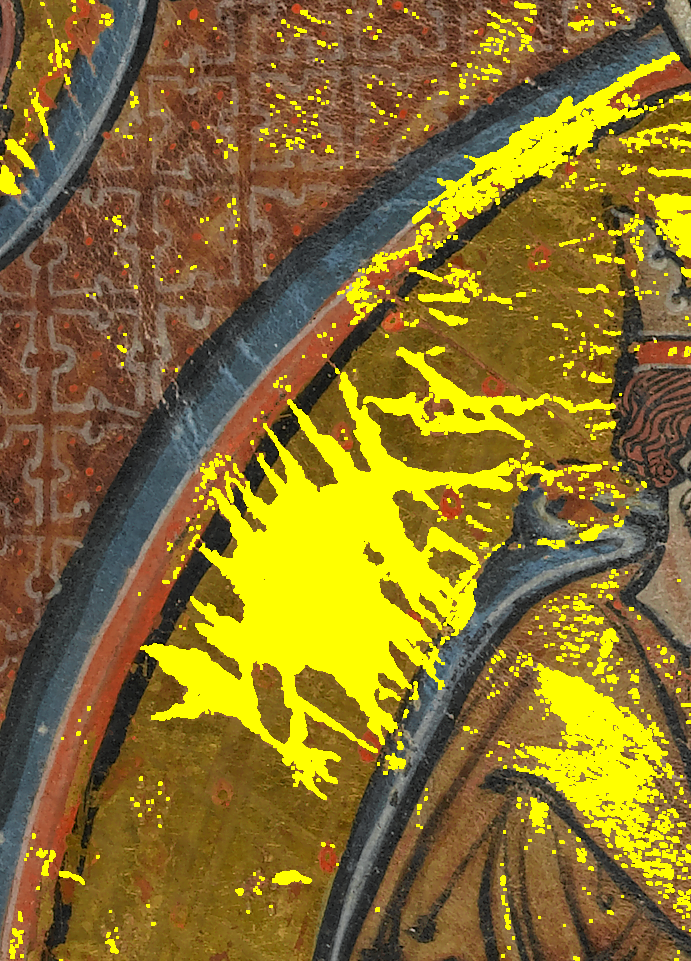}}
\end{center}
\caption{Detail from Figure \ref{fig:chan vese inputs 2}.}
\label{fig:k means segm 2}
  \end{subfigure}
  \caption{\csentence{Global inpainting domain segmentation via $k$-means clustering.}
    The $k$-means clustering algorithm is run on the whole image selection in terms of intensity-based image features. Once the labels are assigned for each pixel, the outputs of the binary classification algorithm showed in Figure \ref{fig:chan vese inputs} and \ref{fig:chan vese inputs 2} are used to compare cluster classes and the extract the ones best fitting the training ones.}
      \label{fig:k means results}
      \end{figure}

                   \begin{figure}[h!]  
  \begin{subfigure}{0.31\textwidth}
\begin{center}
\includegraphics[height=4cm]{inpainting/input203.png}
\end{center}
  \end{subfigure}
      \begin{subfigure}{0.31\textwidth}
\begin{center}
\includegraphics[height=4cm]{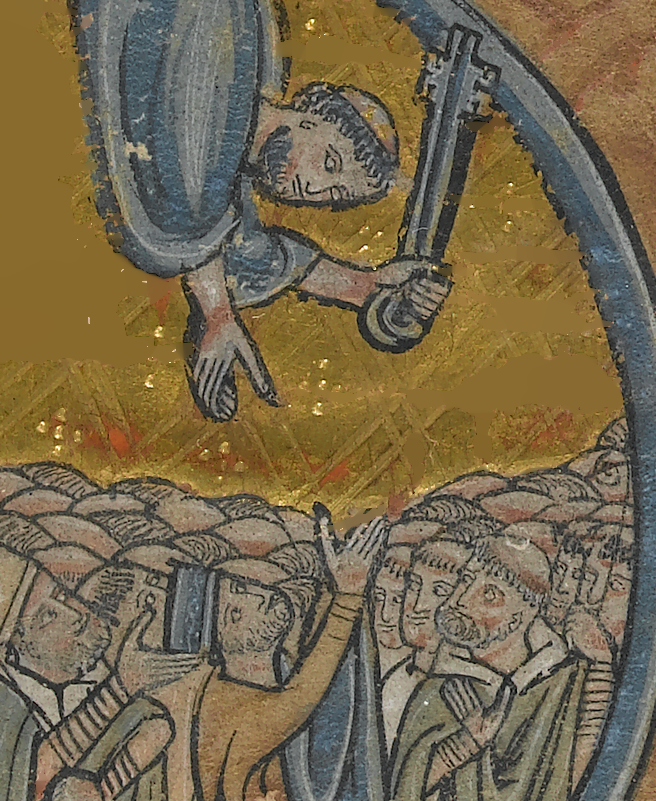}
\end{center}
  \end{subfigure}
        \begin{subfigure}{0.31\textwidth}
\begin{center}
\includegraphics[height=4cm]{\detokenize{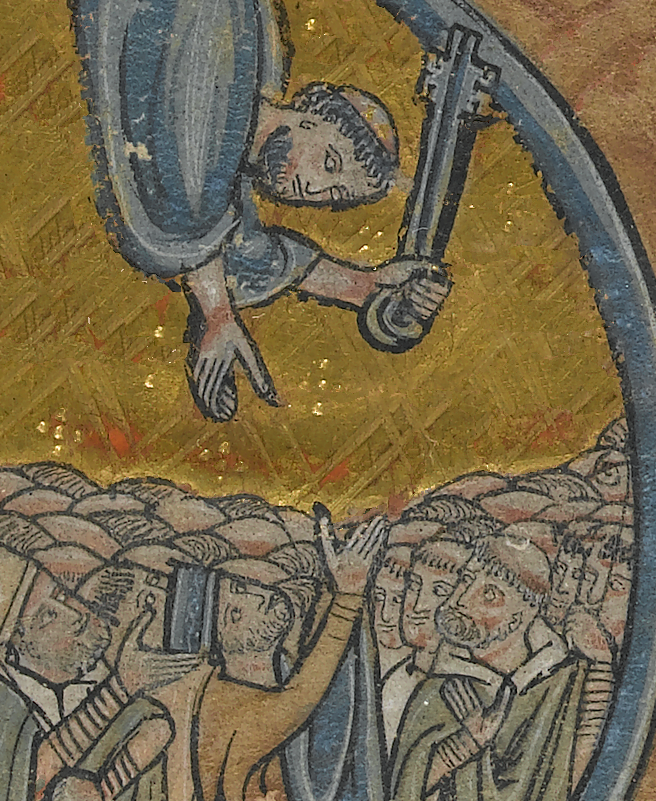}}
\end{center}
  \end{subfigure} \\ \vspace{0.3cm}
  \begin{subfigure}{0.31\textwidth}
\begin{center}
\includegraphics[height=4cm]{inpainting/input101.png}
\end{center}
\caption{Details}
  \end{subfigure}
      \begin{subfigure}{0.31\textwidth}
\begin{center}
\includegraphics[height=4cm]{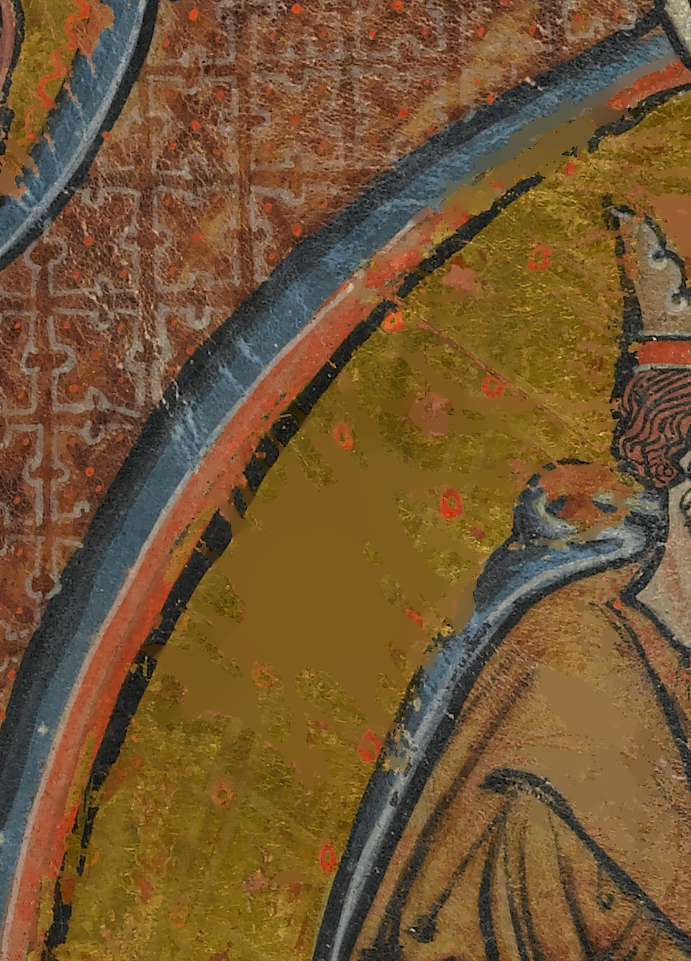}
\end{center}
\caption{TV initialisations}
  \end{subfigure}
        \begin{subfigure}{0.31\textwidth}
\begin{center}
\includegraphics[height=4cm]{\detokenize{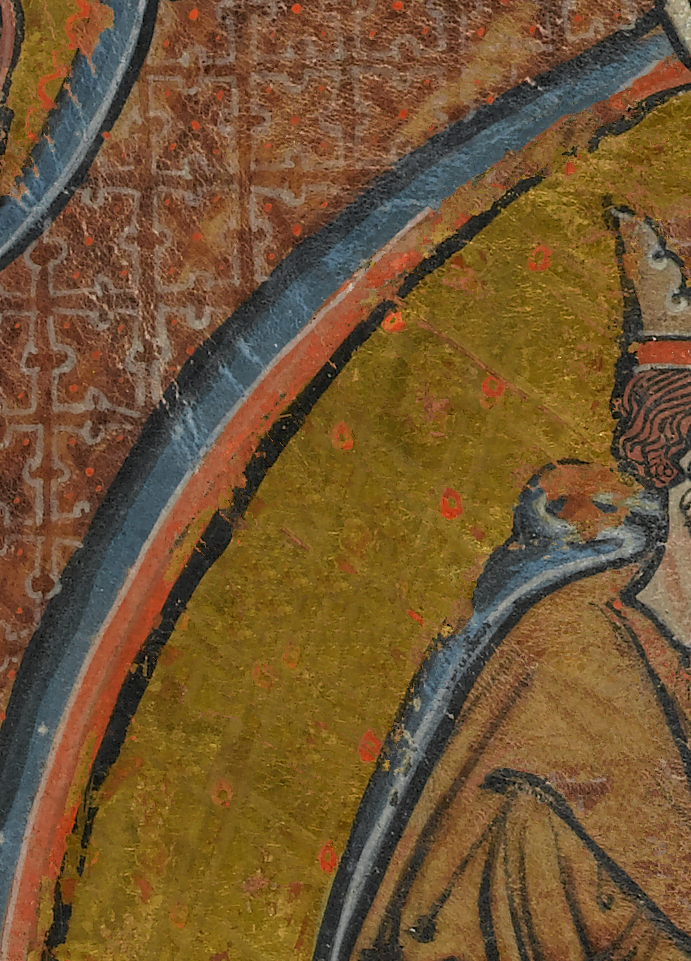}}
\end{center}
\caption{Final results}
  \end{subfigure}
  \caption{\csentence{Inpainting of damaged areas in Figure \ref{fig:chan vese inputs} and  \ref{fig:chan vese inputs 2}.}
     Once the inpainting domain is detected, the TV inpainting model \eqref{eq:functional}-\eqref{eq:TV} is used to provide a good initialisation for the exemplar-based model \eqref{eq:nonlocal inp}. The final result shows the desired transfer of both geometric and texture information in the damaged areas. Patch size: $5\times 5$ (upper row), $7\times 7$ (bottom row).}
      \label{fig:inpainting}
      \end{figure}
      
                \begin{figure}[h!]  
  \begin{subfigure}{0.31\textwidth}
\begin{center}
\includegraphics[height=4cm]{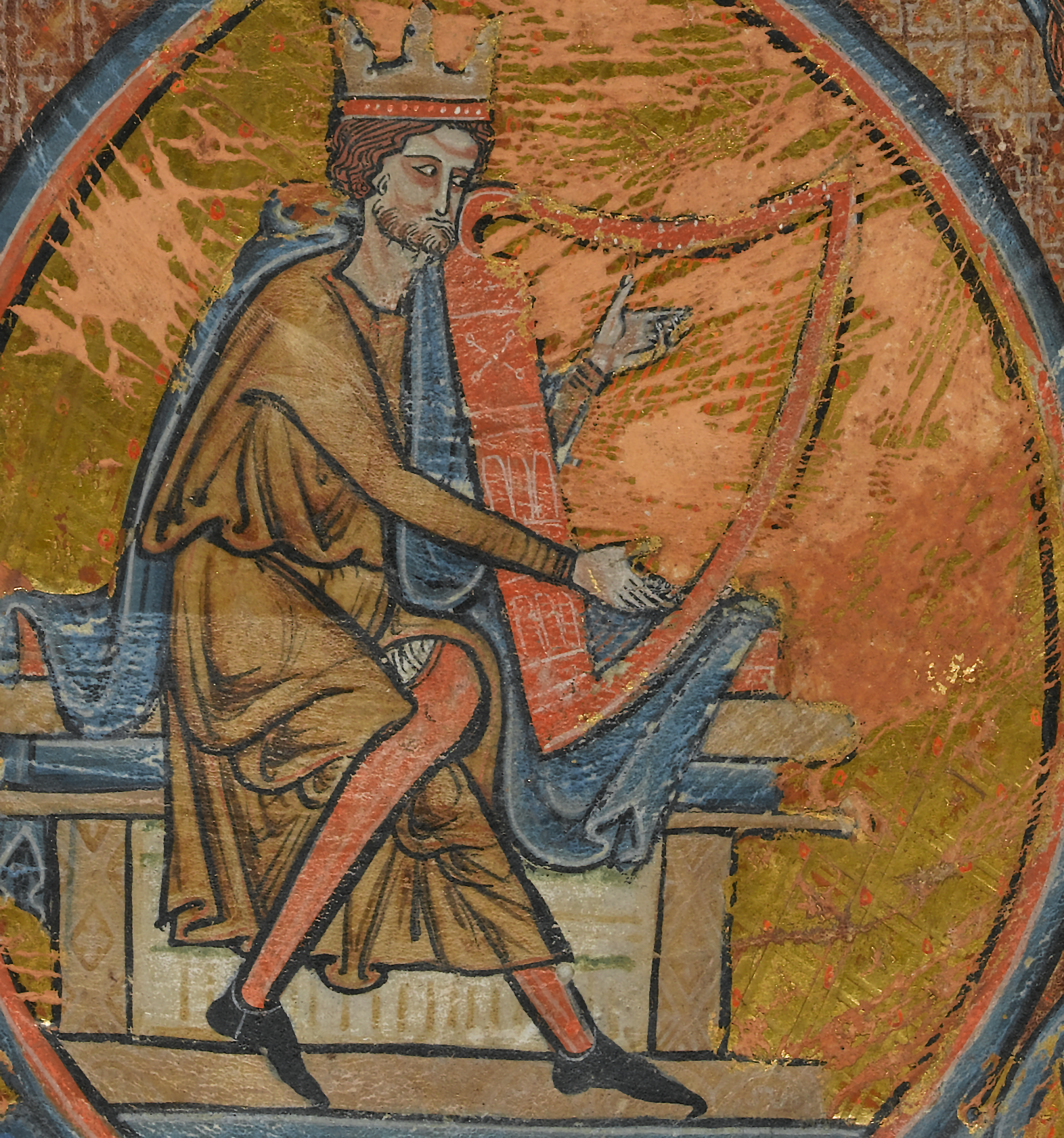}
\end{center}
\caption{Detail.}
  \end{subfigure}
    \begin{subfigure}{0.31\textwidth}
\begin{center}
\includegraphics[height=4cm]{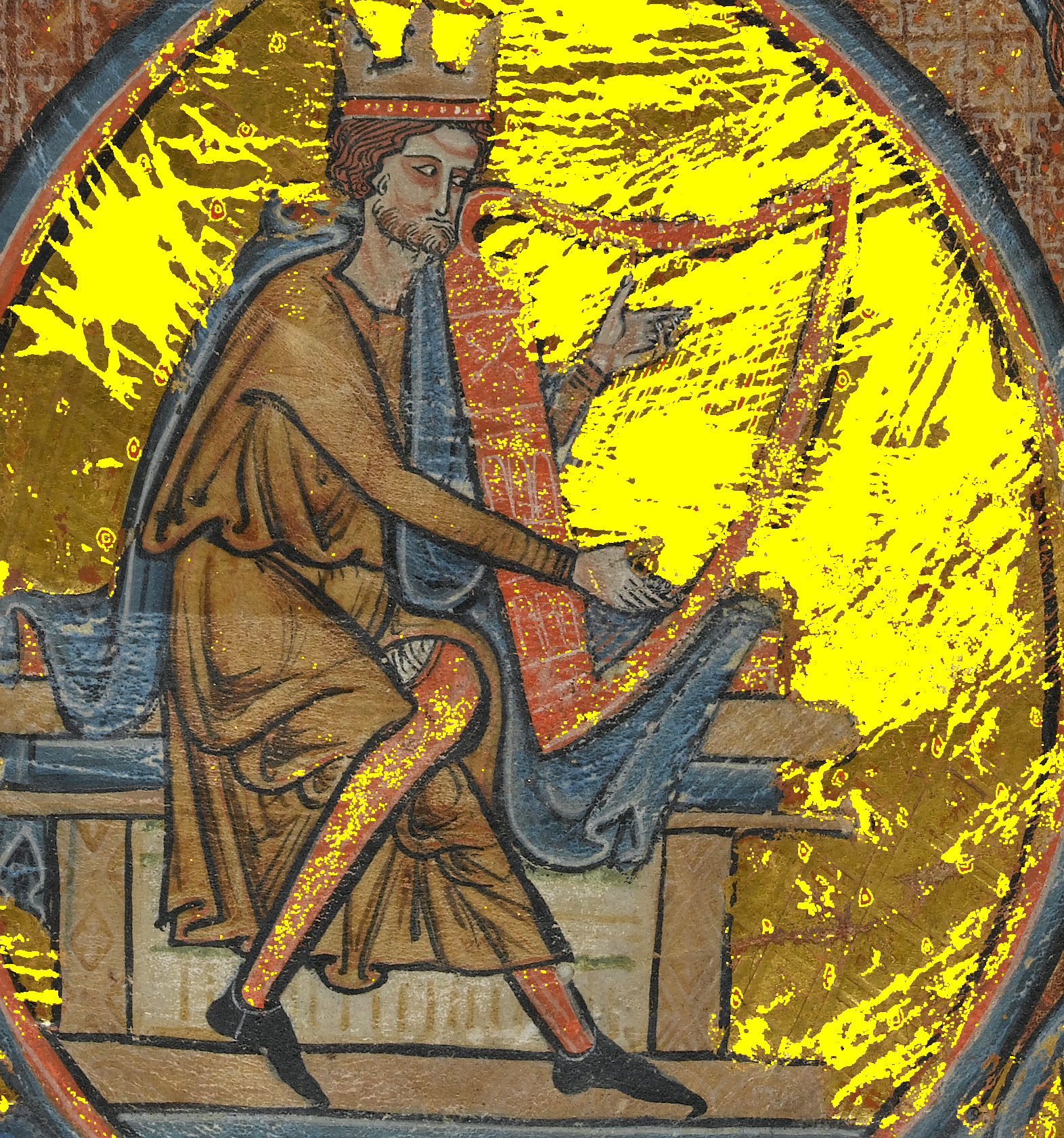}
\end{center}
\caption{Inpainting domain.}
  \end{subfigure}
      \begin{subfigure}{0.31\textwidth}
\begin{center}
\includegraphics[height=4cm]{\detokenize{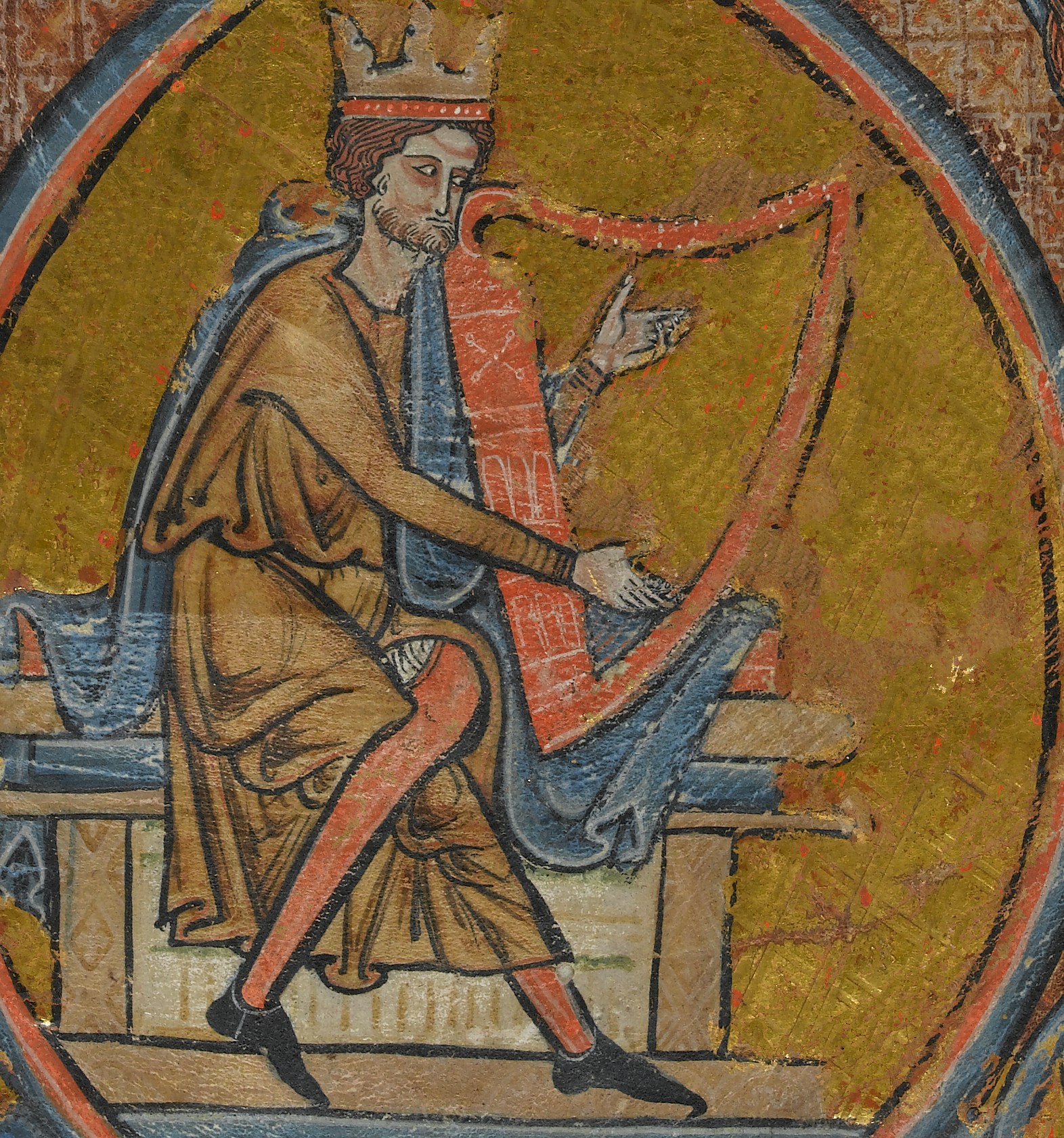}}
\end{center}
\caption{Final result.}
  \end{subfigure}
  \caption{\csentence{Inpainting of large image region with large damaged areas.}
    Inpainting results of the combined model for a large detail ($1572\times 1681$ pixels) with large damaged areas. Patch size: $9\times 9$.}
      \label{fig:inpainting 2}
      \end{figure}
      
       \begin{figure}[h!] \label{fig:workflow segmentation}
  \includegraphics[height=7cm]{\detokenize{inpainting/diagram_segmentation}}
  \caption{\csentence{Workflow of the combined algorithm for inpainting.}
     The diagram describes the different steps of the combined semi-supervised algorithm for inpainting domain detection followed by the restoration of the damaged areas via mathematical inpainting. Boxes requiring user inputs are coloured \emph{orange}, where the ones where automatic steps are performed are coloured \emph{blue}. The final objective is coloured \emph{green}.}
      \end{figure}
      
\begin{figure}[h!]
    \includegraphics[width=5cm]{\detokenize{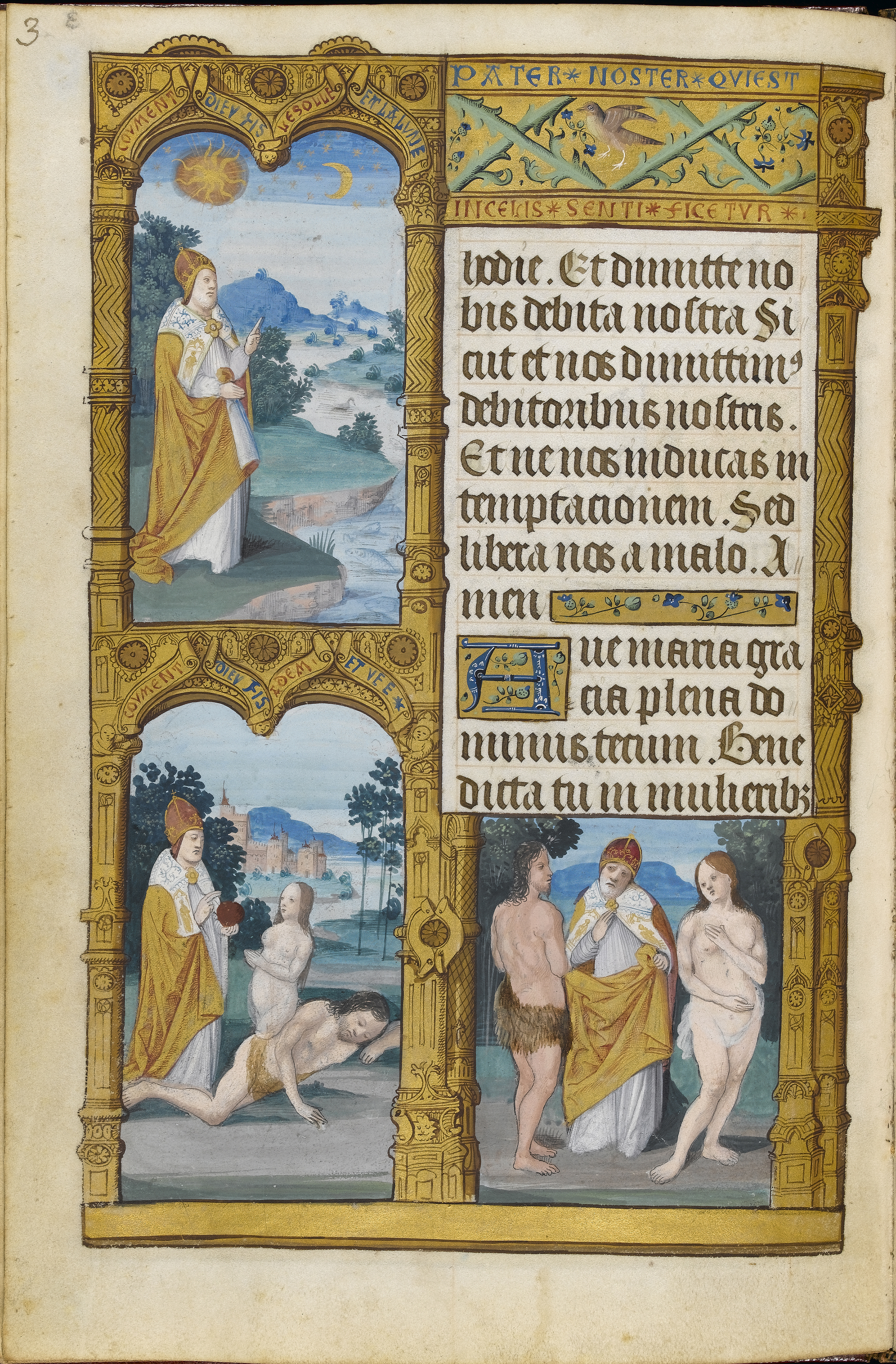}} \qquad
    \includegraphics[width=5cm]{\detokenize{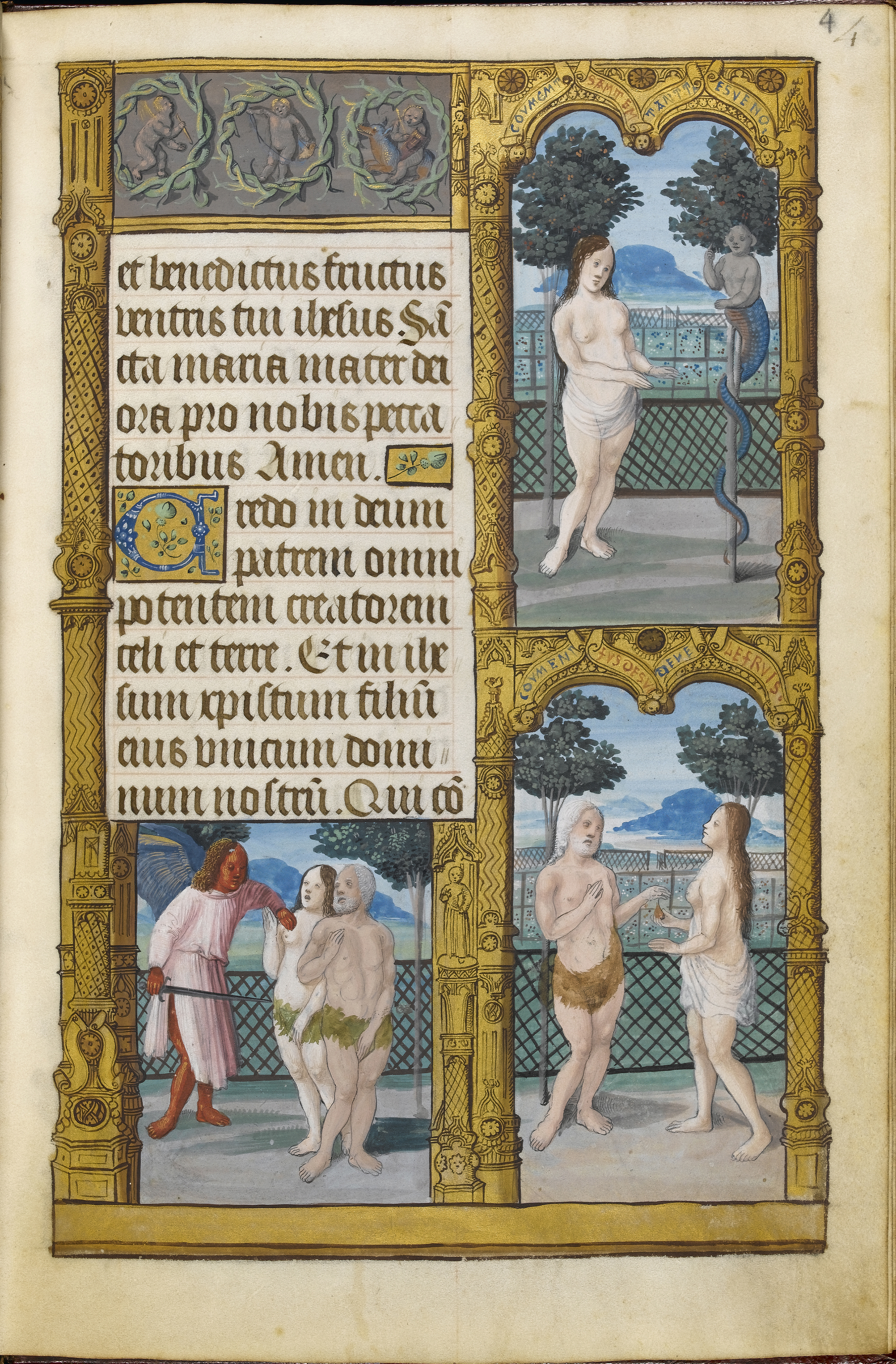}}
    \caption{\csentence{Illuminated manuscripts}
    Two illuminated manuscripts touched up to cover Adam and Eve's nudity, see Section \ref{sec:osmosis} for more details.}
    \label{fig:osmosisdata}
\end{figure}

\begin{figure}[h!]
    \begin{subfigure}{0.45\textwidth}
        \begin{center}
            \includegraphics[width=0.9\linewidth]{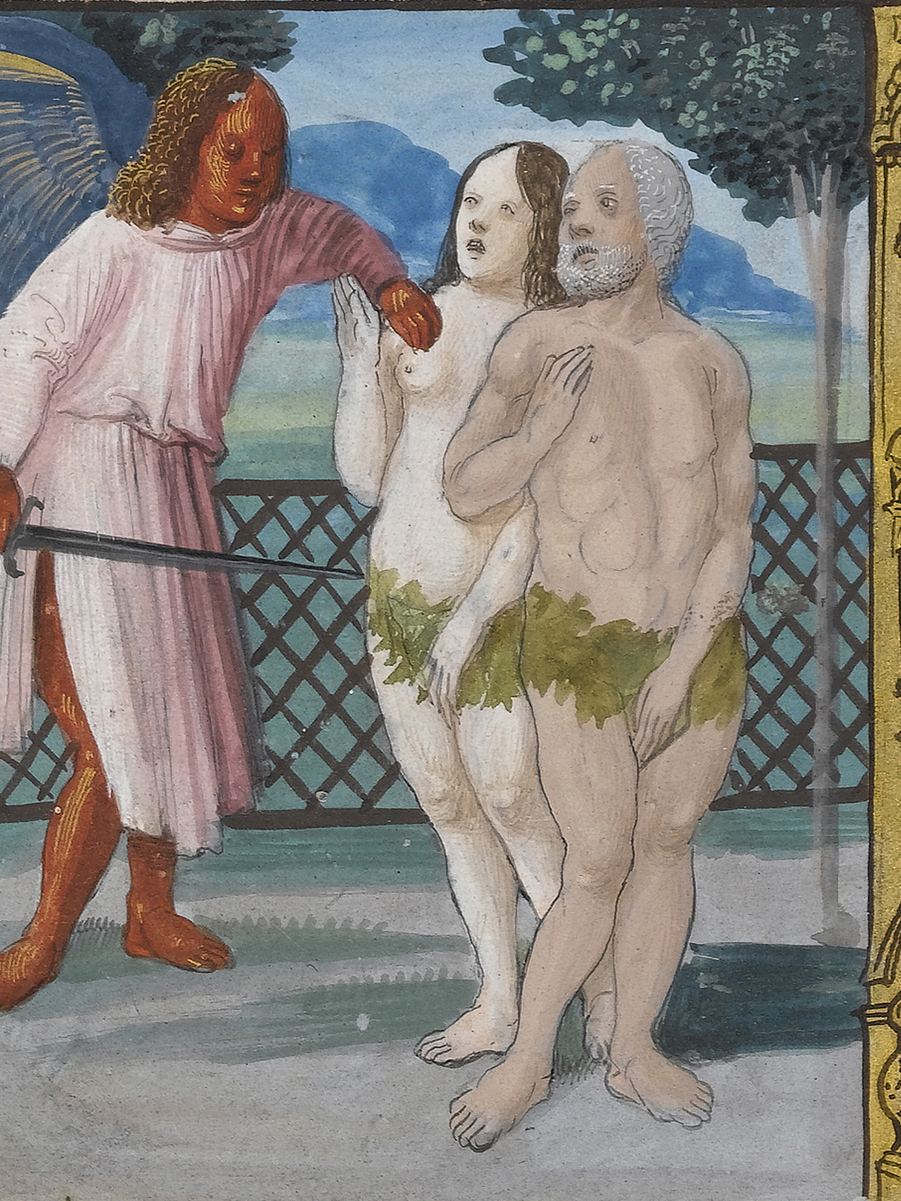}
        \end{center}
        \caption{RGB}
    \end{subfigure}
    \begin{subfigure}{0.45\textwidth}
        \begin{center}
            \includegraphics[width=0.9\linewidth]{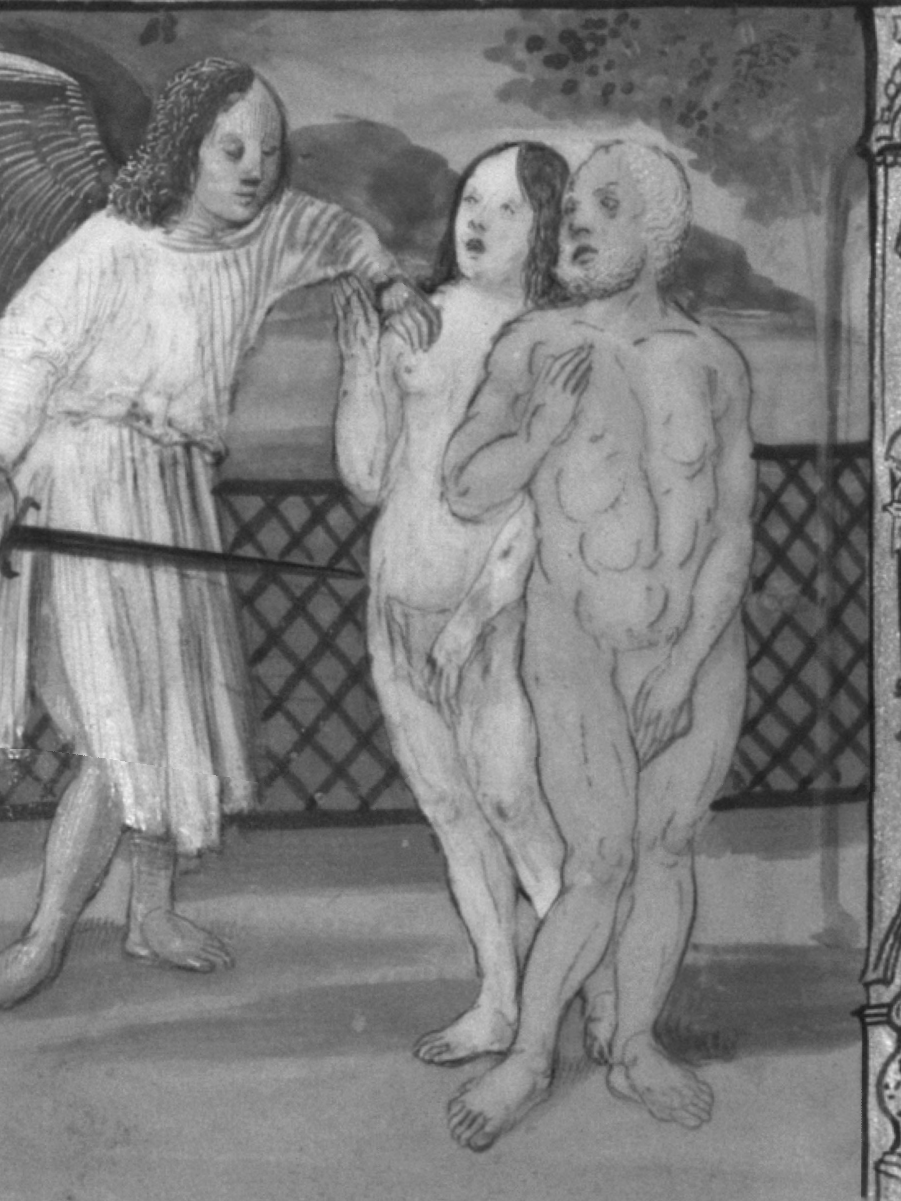}
        \end{center}
        \caption{Infrared}
    \end{subfigure}
    \begin{subfigure}{0.45\textwidth}
        \begin{center}
            \includegraphics[width=0.9\linewidth]{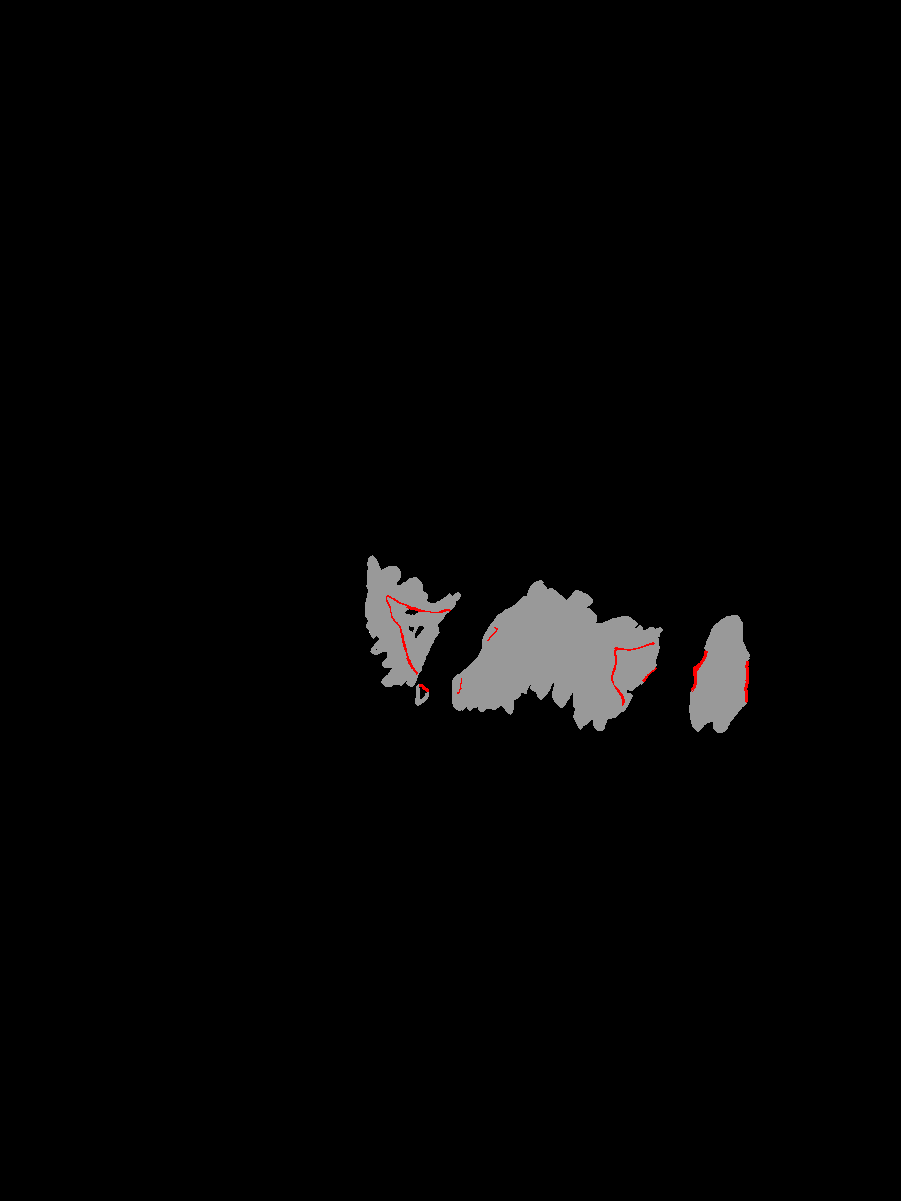}
        \end{center}
        \caption{Mask}
    \end{subfigure}
    \begin{subfigure}{0.45\textwidth}
        \begin{center}
            \includegraphics[width=0.9\linewidth]{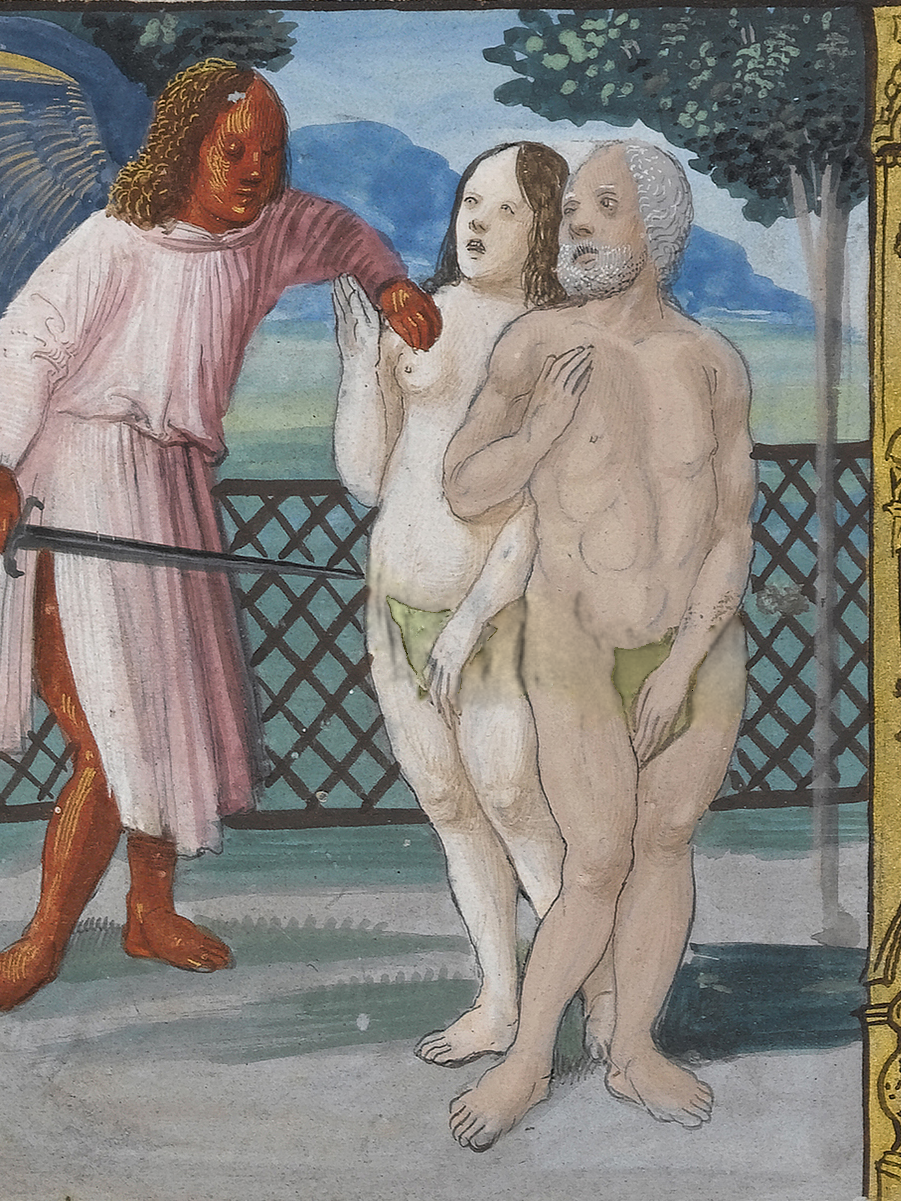}
        \end{center}
        \caption{Final Result}
    \end{subfigure}
    \caption{\csentence{The added paint doesn't appear on the inrared}
    It is a simple case of seamless cloning on the subdomain marked in gray on the mask with some Neumann dirichlet boundary conditions along edges separating the skin and green areas (background and fig leaves), encoded by the red lines. The final result shows for Eve some undesirable dark lines present in the infrared.
                }
                \label{fig:adameveleft}
\end{figure}

\begin{figure}[h!]
    \begin{subfigure}{0.45\textwidth}
        \begin{center}
            \includegraphics[width=0.9\linewidth]{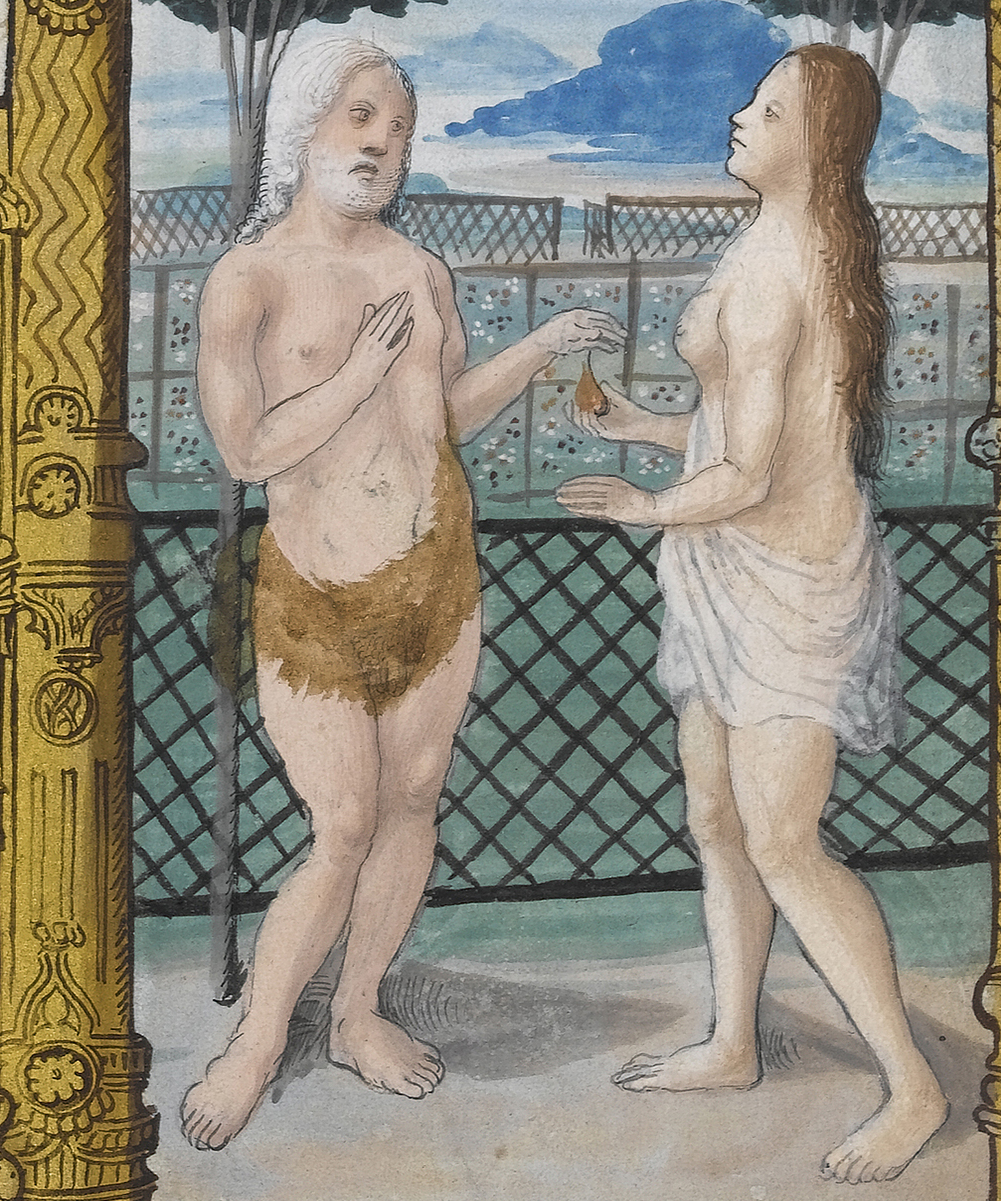}
        \end{center}
        \caption{RGB}
    \end{subfigure}
    \begin{subfigure}{0.45\textwidth}
        \begin{center}
            \includegraphics[width=0.9\linewidth]{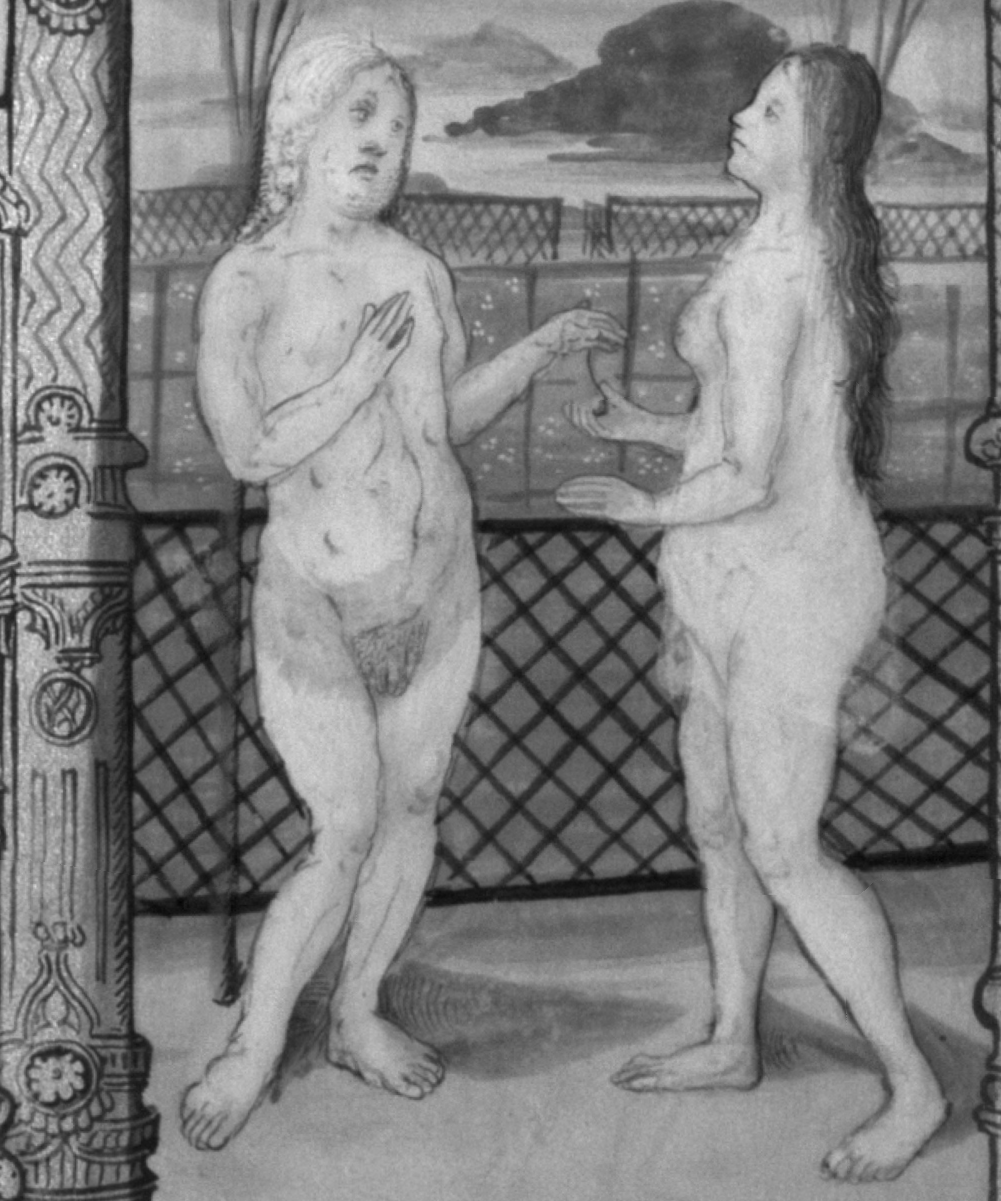}
        \end{center}
        \caption{Infrared}
    \end{subfigure}
    \begin{subfigure}{0.45\textwidth}
        \begin{center}
            \includegraphics[width=0.9\linewidth]{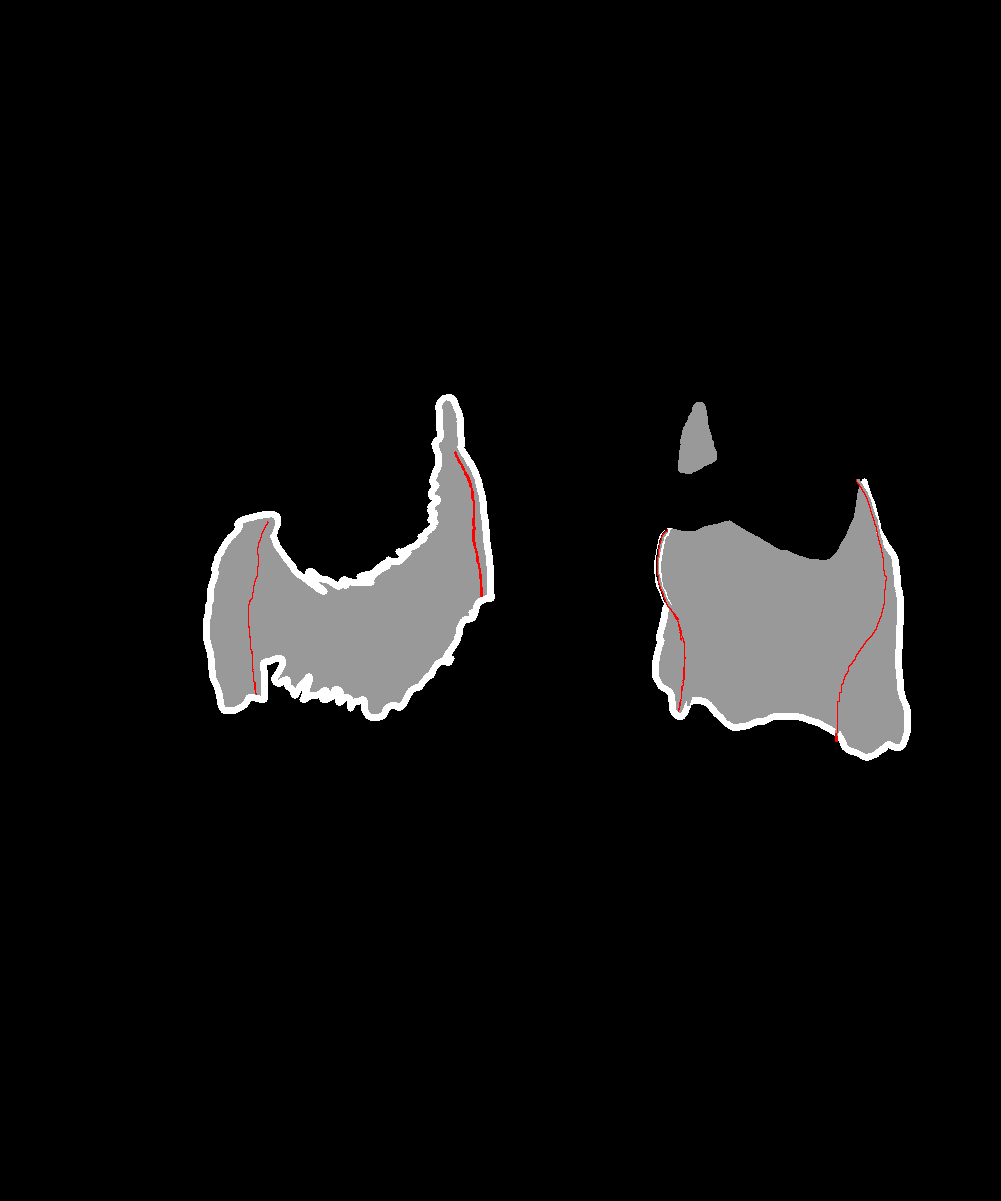}
        \end{center}
        \caption{Mask}
    \end{subfigure}
    \begin{subfigure}{0.45\textwidth}
        \begin{center}
            \includegraphics[width=0.9\linewidth]{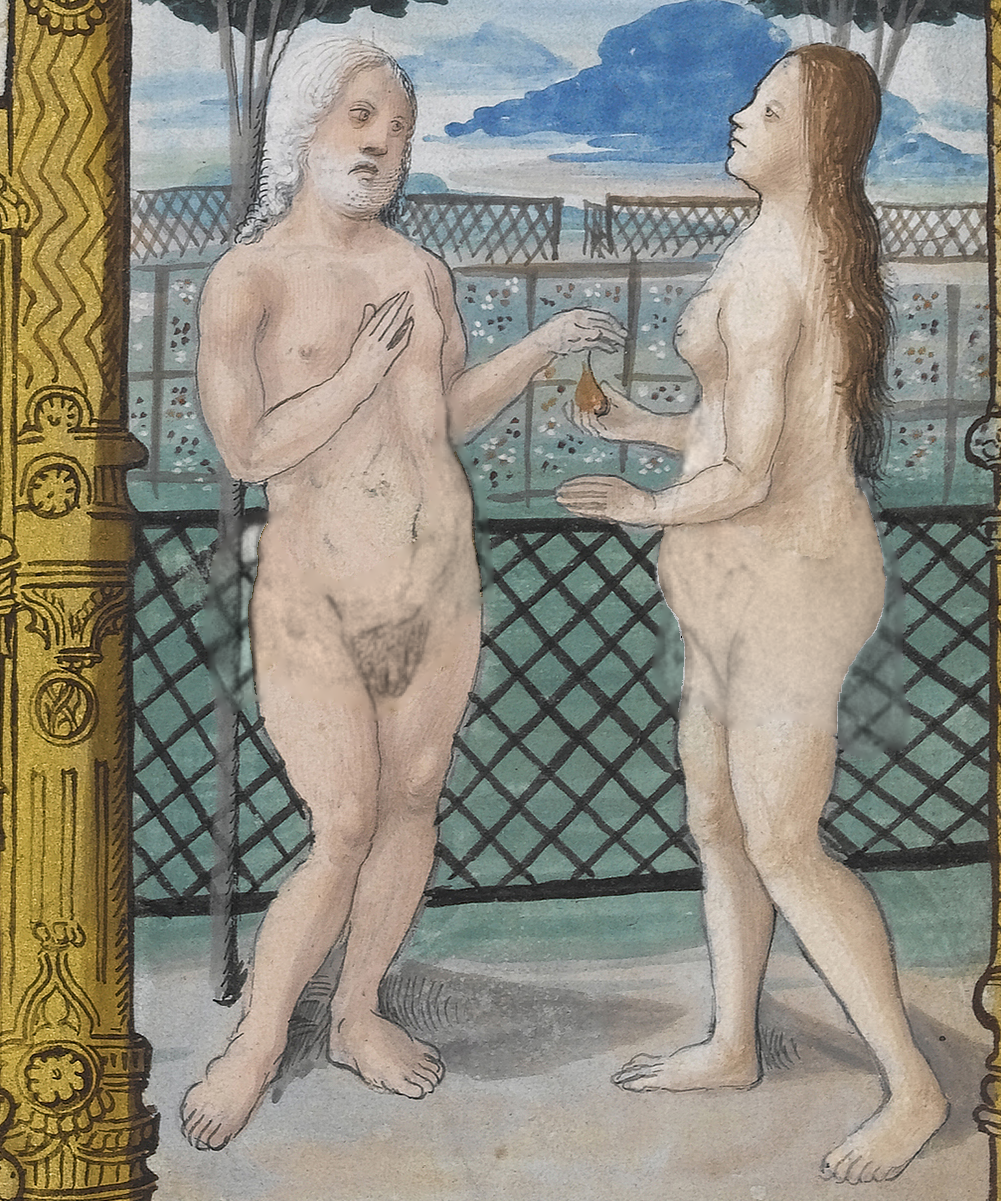}
        \end{center}
        \caption{Final Result}
    \end{subfigure}
    \caption{\csentence{The added paint appears on the infrared}
    Because the edge of the added cloth appears on the infrared, we put the drift-field to zero before solving the equation (white lines in the mask). Some texture from the added cloth that appears on the infrared is still noticeable in the final results.
    } \label{fig:adameveright}
\end{figure}

\begin{figure}[h!]
    \begin{subfigure}{0.31\textwidth}
        \begin{center}
            \includegraphics[width=0.9\linewidth]{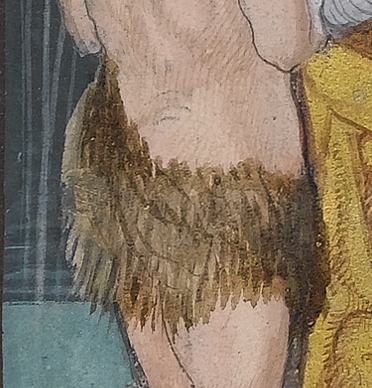}
        \end{center}
        \caption{RGB}
    \end{subfigure}
    \begin{subfigure}{0.31\textwidth}
        \begin{center}
            \includegraphics[width=0.9\linewidth]{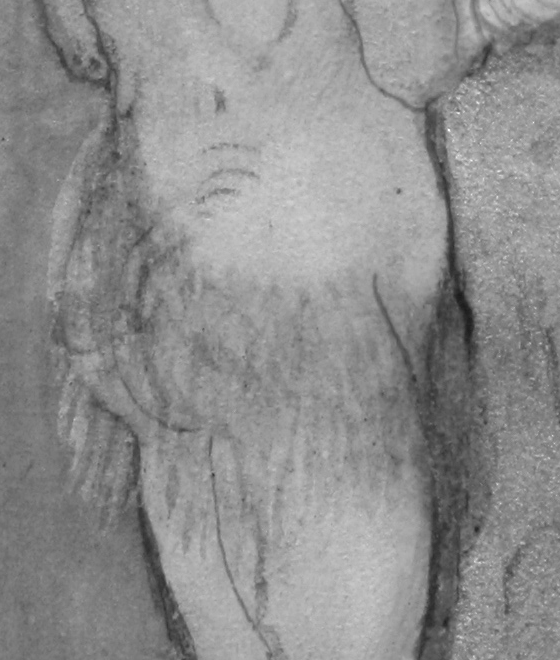}
        \end{center}
        \caption{Infrared}
    \end{subfigure}
    \begin{subfigure}{0.31\textwidth}
        \begin{center}
            \includegraphics[width=0.9\linewidth]{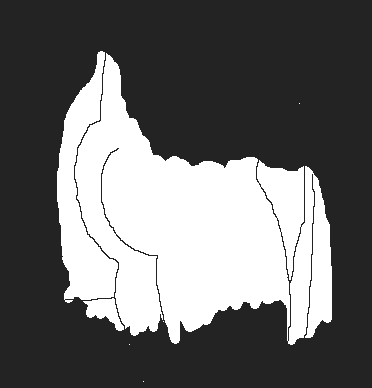}
        \end{center}
        \caption{Handdrawn sketch}
    \end{subfigure}
    \begin{subfigure}{0.31\textwidth}
        \begin{center}
            \includegraphics[width=0.9\linewidth]{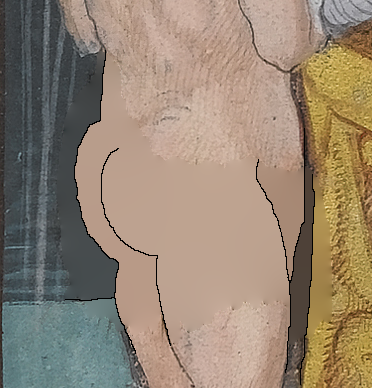}
        \end{center}
        \caption{After seamless cloning}
    \end{subfigure}
    \begin{subfigure}{0.31\textwidth}
        \begin{center}
            \includegraphics[width=0.9\linewidth]{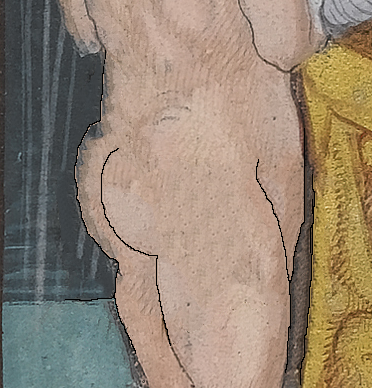}
        \end{center}
        \caption{Final Result}
    \end{subfigure}
    \caption{\csentence{The added cloth texture is on the infrared}
    As too much texture from the added cloth is on the infrared, we have no other solution than performing a seamless cloning step using a hand drawn sketch. This however makes all texture disappear in the restored area. We therefore add an exemplar-based inpainting step to recover a more natural looking output.
    } \label{fig:adamleft}
\end{figure}

\begin{figure}[h!]
\begin{subfigure}{.45\linewidth}
\begin{center}
\includegraphics[height=7cm]{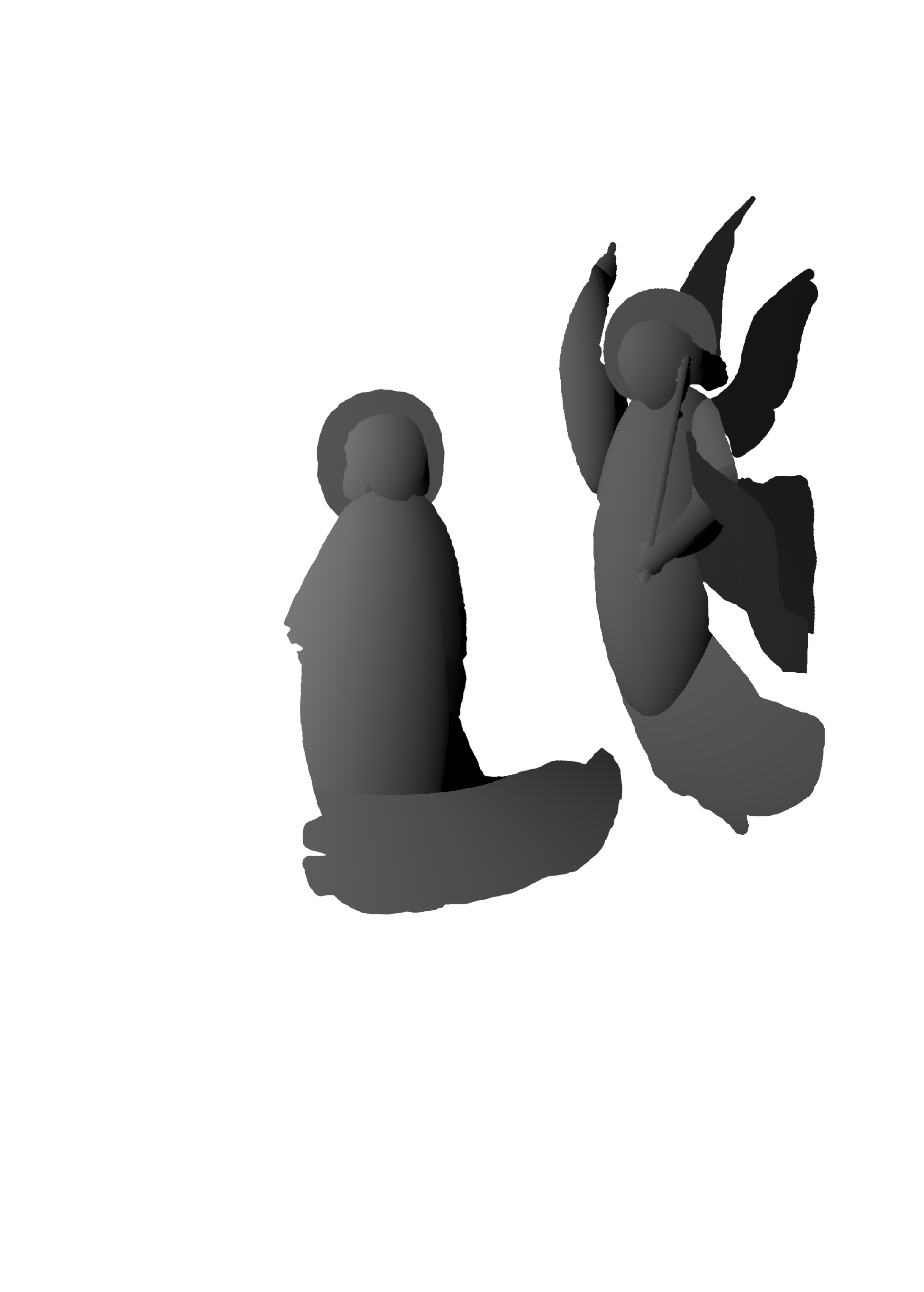}
\end{center}
\caption{3D models of the virgin Mary and angel Gabriel, after clipping.}
\end{subfigure}
\begin{subfigure}{.45\linewidth}
\begin{center}
\includegraphics[height=7cm]{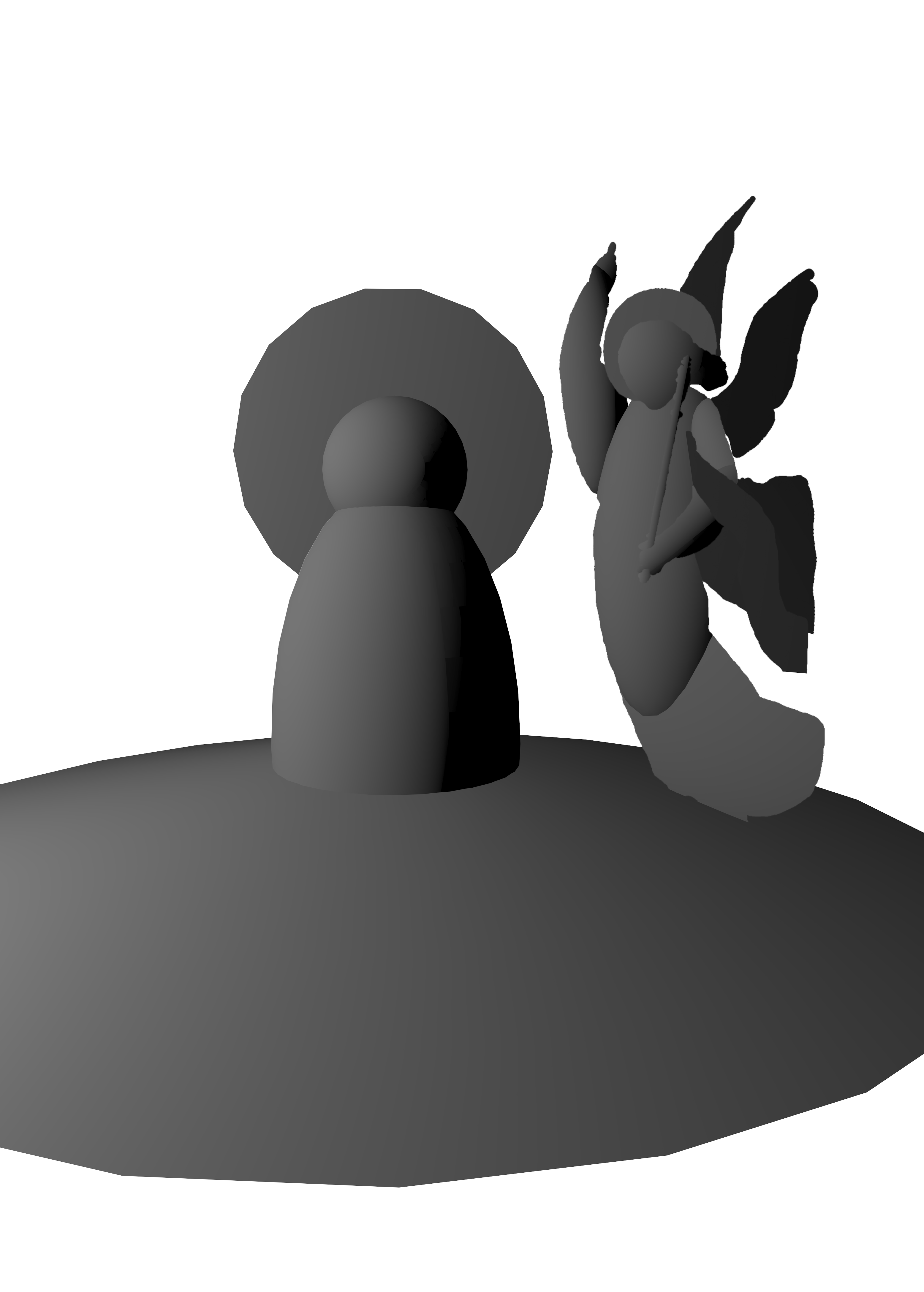}
\end{center}
\caption{3D models of the angel Gabriel and the virgin Mary.  This time the geometry of Mary is unclipped.}
\end{subfigure}
\caption{{\bf 3D models used in the conversion pipeline.}  Illustration of some of the models used for the conversion of the manuscript in Figure \ref{fig:twoEyes}.  When clipping is turned off, we see that the rough geometry for the virgin Mary generated in step 1 of the pipeline is, in fact, just a couple of simple geometric shapes.}
\label{fig:geometry}
\end{figure}

\begin{figure}[h!]
\begin{subfigure}{.45\linewidth}
\begin{center}
\includegraphics[height=7cm]{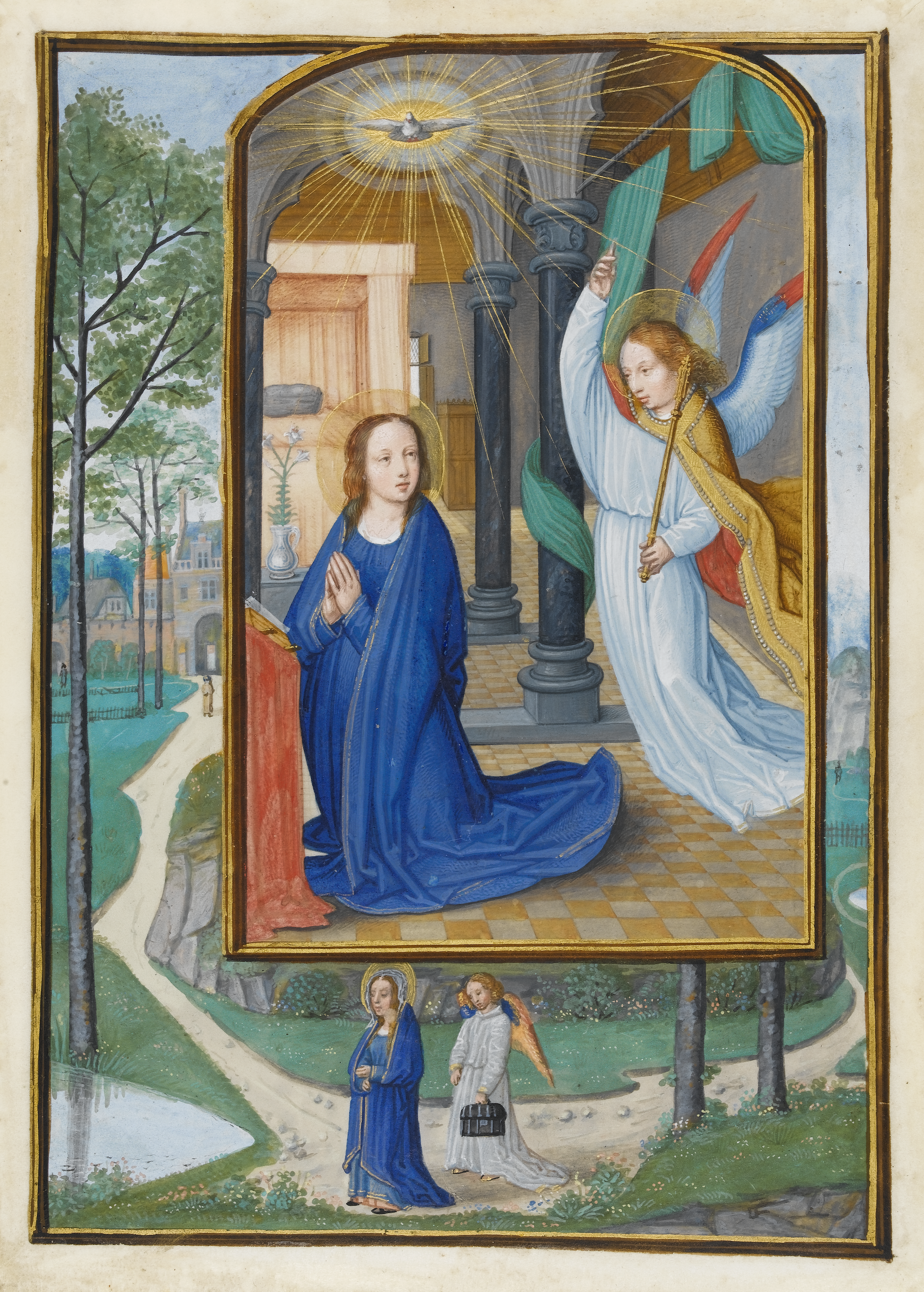}
\end{center}
\caption{Original manuscript (right eye view).}
\end{subfigure}
\begin{subfigure}{.45\linewidth}
\begin{center}
\includegraphics[height=7cm]{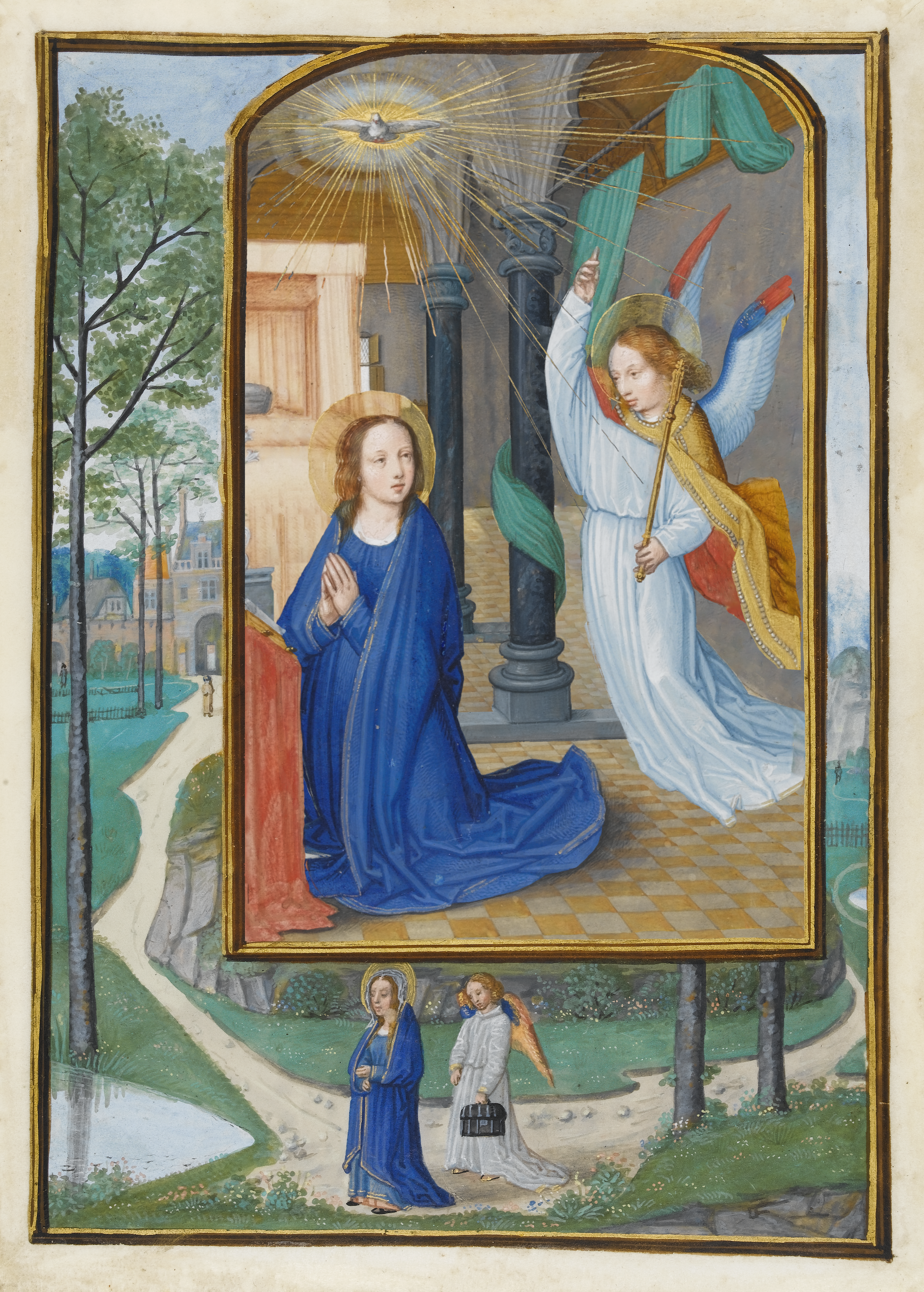}
\end{center}
\caption{Reconstructed left eye view.}
\end{subfigure}
\caption{{\bf 3D conversion of an illuminated manuscript.}  The illuminated manuscript considered here is 
Simon Bening, \emph{Annunciation}, Fitzwilliam Museum, MS 294b, Flanders, Bruges, (1522-1523).
The restored manuscript (a) is converted into a stereo 3D pair.  To view the resulting stereo 3D image without glasses, first cross your eyes so that each image splits in two.  Make the middle two images overlap, and then bring the superimposed image into focus (try varying your distance from the computer screen).}
\label{fig:twoEyes}
\end{figure}

\begin{figure}[h!]
\begin{subfigure}{.4\linewidth}
\begin{center}
\includegraphics[height=4cm]{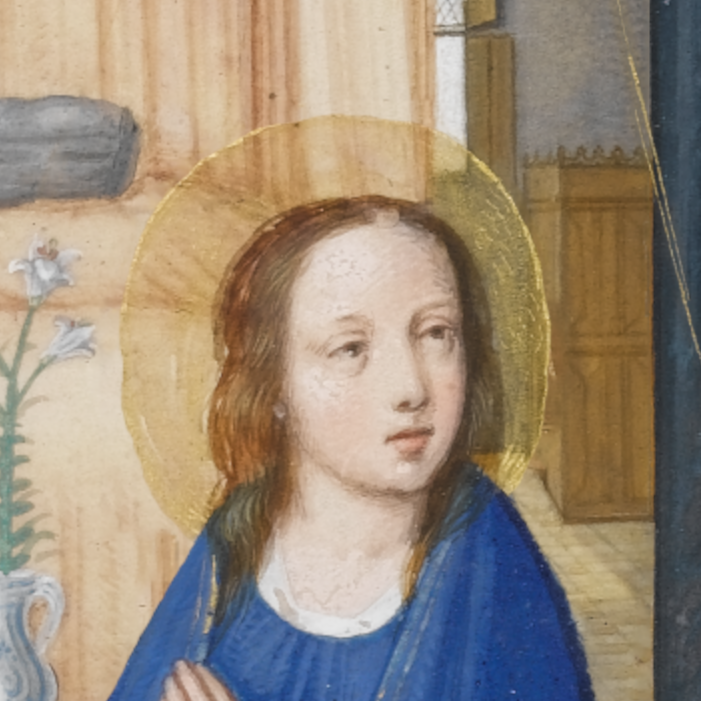}
\end{center}
\caption{Original right eye view.}
\end{subfigure}
\begin{subfigure}{.4\linewidth}
\begin{center}
\includegraphics[height=4cm]{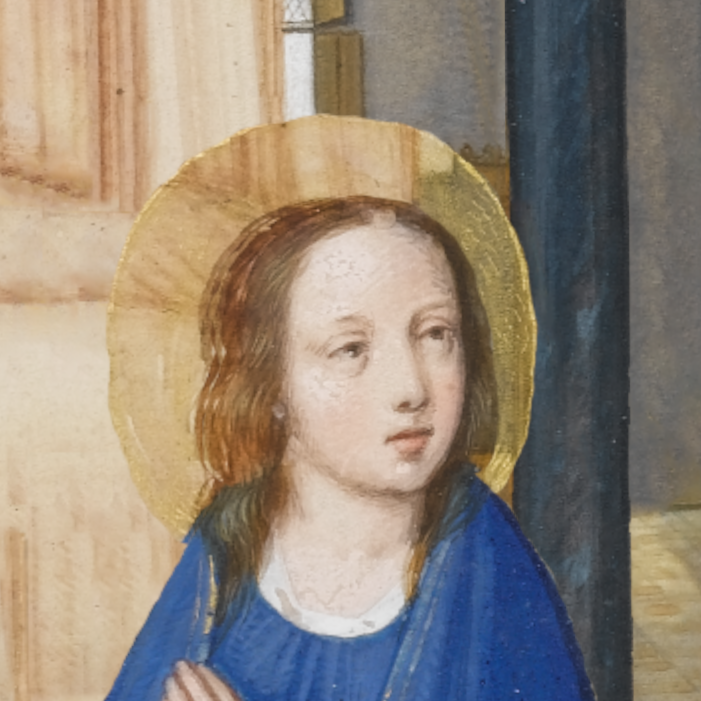}
\end{center}
\caption{Reconstructed left eye view.}
\end{subfigure}
\caption{{\bf Failure for semi-transparent surfaces.}  Closeup of the halo of the virgin Mary in the original right eye view (a) and the reconstructed left eye view (b).  In (a), we are able to see part of the background - in this case Mary's bed - through her halo.  In (b), the same chunk of background is incorrectly carried over to the new location, obscuring the actual background.}
\label{fig:limitation}
\end{figure}

\begin{figure}[h!]
\begin{subfigure}{.2\linewidth}
\includegraphics[height=4cm]{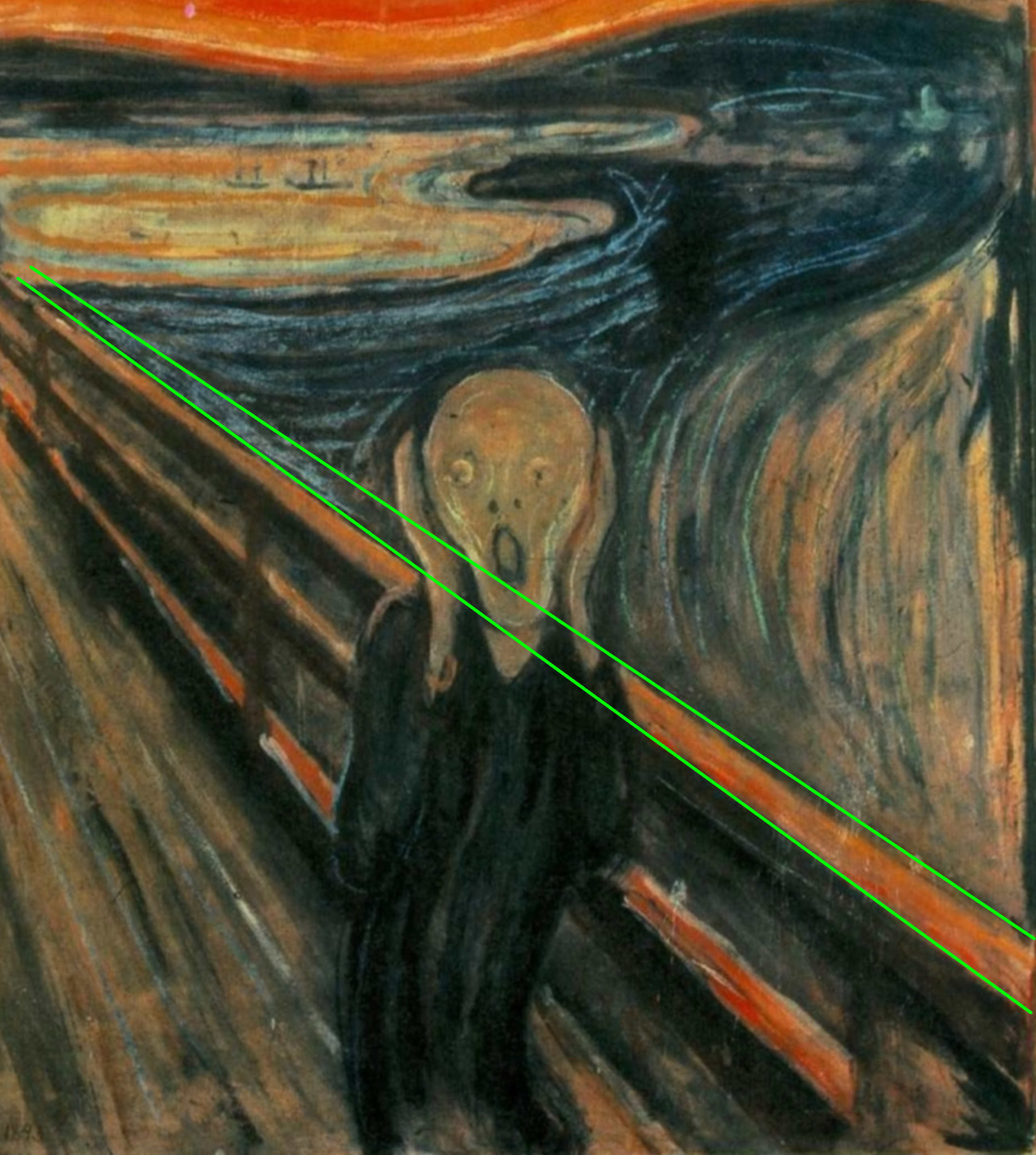}
\caption{Detail from {\em The Scream} with superimposed lines.}
\end{subfigure}
\hfill
\begin{subfigure}{.5\linewidth}
\includegraphics[height=4cm]{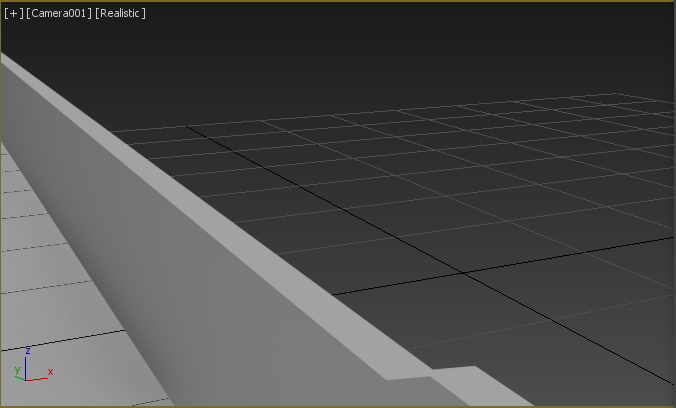}
\caption{3D reconstruction of bridge with a special ``kink'' added.}
\end{subfigure} 
\caption{{\bf The Scream.}  As another example, we applied the 3D conversion pipeline from Section \ref{sec:pipeline} to Edvard Munch's painting {\em The Scream}.  In the process we discovered, as in (a), that the railing of the bridge in the painting does not obey the laws of perspective.  To get around this issue, we had to introduce a ``kink'' into our 3D model of the bridge, as in (b).}
\label{fig:scream}
\end{figure}







\end{backmatter}
\end{document}